\documentclass[letterpaper, 10 pt, journal, twoside]{IEEEtran}

\usepackage{amsmath}
\usepackage{amssymb}
\usepackage{booktabs}
\usepackage{caption}
\usepackage{subfigure}
\usepackage[dvips]{graphicx}
\usepackage{multirow}
\usepackage{amsfonts}
\usepackage{breakurl}
\usepackage{enumerate}
\usepackage{tabularx}
\usepackage{algorithm,algorithmic}
\usepackage{bm}
\usepackage{enumerate}
\usepackage{adjustbox}
\usepackage{xcolor}
\usepackage{siunitx}
\usepackage{booktabs}
\usepackage{mdframed}
\usepackage{adjustbox}
\usepackage{authblk}
\usepackage{color}
\usepackage{xurl}
\usepackage{cite}
\makeatletter
\let\NAT@parse\undefined
\makeatother
\usepackage{float}
\usepackage{pifont}
\usepackage{hyperref}
\usepackage{amssymb}
\usepackage{mathtools}

\newcommand{\prths}[1]{\left(#1\right)}

\newcommand{\vecg}{\mathbf{g}}

\newcommand{\vecx}{\mathbf{x}}

\newcommand{\vecu}{\mathbf{u}}

\newcommand{\vecxi}{\bm{\xi}}

\newcommand{\matM}{\mathbf{M}}

\newcommand{\matR}{\bm{\mathit{R}}}

\newcommand{\inertia}{\mathbf{J}}

\newcommand{\loadmass}{m_{L}}

\newcommand{\angvel}{\mathbf{\Omega}}
\newcommand{\angacc}{\dot{\angvel}}

\newcommand{\robotpos}{\vecx_{Q}}
\newcommand{\robotrot}{\matR_{Q}}

\newcommand{\robotvel}{\dot{\vecx}_{Q}}
\newcommand{\robotacc}{\ddot{\vecx}_{Q}}
\newcommand{\robotangvel}{\angvel}
\newcommand{\robotangacc}{\angacc}

\newcommand{\rot}[1]{\matR_{#1}}

\newcommand{\loadpos}{\vecx_{L}}

\newcommand{\loadvel}{\dot{\vecx}_{L}}

\newcommand{\loadacc}{\ddot{\vecx}_{L}}

\newcommand{\loadjerk}{\dddot{\vecx}_{L}}

\newcommand{\ybf}[1]{\mathbf{y}_{#1}}

\newcommand{\cablevec}[1]{\vecxi_{#1}}

\newcommand{\cabledotvec}[1]{\dot{\vecxi}_{#1}}

\newcommand{\cableddotvec}[1]{\ddot{\vecxi}_{#1}}

\newcommand{\realnum}[1]{\mathbb{R}^{#1}}
\newcommand{\SOthree}{SO(3)}

\newcommand{\worldf}{\mathcal{I}}
\newcommand{\robotf}{\mathcal{B}}

\newcommand{\axis}[2]{\mathbf{e}_{#1}^{#2}}

\title{
PolyFly: Polytopic Optimal Planning for Collision-Free Cable-Suspended Aerial Payload Transportation
}

\author{Mrunal Sarvaiya$^{1}$, Guanrui Li$^{2}$, and Giuseppe Loianno$^{1}$
\thanks{Manuscript received: August 13, 2025; Revised: November 12, 2025; Accepted: December 29, 2025.}
\thanks{This paper was recommended for publication by Editor Soon-Jo Chung upon evaluation of the Associate Editor and Reviewers' comments.}
\thanks{This work was supported by the NSF CPS Grant CNS-2603416, the NSF CAREER Award 2546659, and the DARPA YFA Grant D22AP00156-00.}
\thanks{
$^1$The authors are with the University of California Berkeley,
Department of Electrical Engineering and Computer Sciences,
Berkeley, CA 94720, USA. {\tt\footnotesize email: \{mrunaljsarvaiya, loiannog\}@eecs.berkeley.edu}.}
\thanks{$^2$The author is with the Worcester Polytechnic Institute, Robotics Engineering, Worcester, MA 01609, USA. {\tt\footnotesize email: gli7@wpi.edu}.}
\thanks{Digital Object Identifier (DOI): see top of this page.}
}

\begin{document}

\markboth{IEEE Robotics and Automation Letters. Preprint Version. Accepted December, 2025}
{Sarvaiya \MakeLowercase{\textit{et al.}}: PolyFly} 
\makeatletter

\g@addto@macro\@maketitle{
\setcounter{figure}{0}
   \centering
    \includegraphics[width=\textwidth, trim=70 260 150 100, clip]{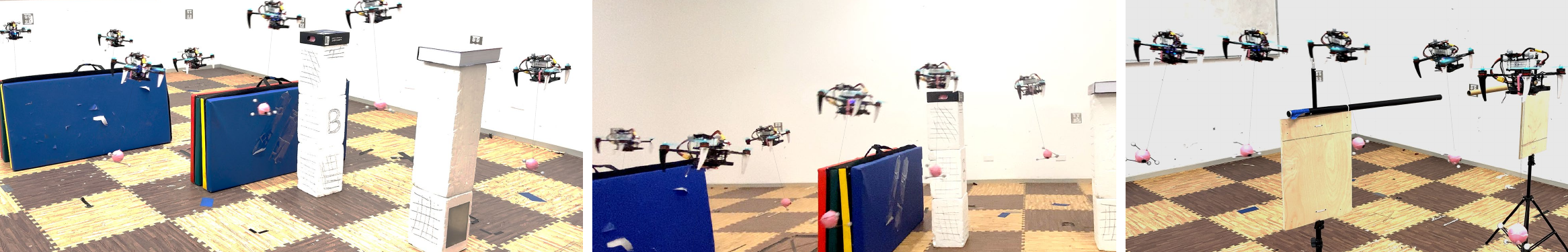} 
    \vspace{-10pt}
    \captionof{figure}{A quadrotor with a suspended payload navigating through cluttered environments. The left and middle images show different perspectives of Env. 3, where the system maneuvers around multiple obstacles. The right image depicts Env. 7, where the system successfully passes through a narrow gap - just wide enough for the cable - demonstrating the planner’s ability to exploit the geometry of individual components.}
    \label{fig:cover_page}
    \vspace{-20pt}
}
\makeatother
\maketitle

\begin{abstract}
Aerial transportation robots using suspended cables have emerged as versatile platforms for disaster response and rescue operations. To maximize the capabilities of these systems, robots need to aggressively fly through tightly constrained environments, such as dense forests and structurally unsafe buildings, while minimizing flight time and avoiding obstacles. Existing methods geometrically over-approximate the vehicle and obstacles, leading to conservative maneuvers and increased flight times. We eliminate these restrictions by proposing PolyFly, an optimal global planner which considers a non-conservative representation for aerial transportation by modeling each physical component of the environment, and the robot (quadrotor, cable and payload), as independent polytopes. We further increase the model accuracy by incorporating the attitude of the physical components by constructing orientation-aware polytopes. The resulting optimal control problem is efficiently solved by converting the polytope constraints into smooth differentiable constraints via duality theory. We compare our method against the existing state-of-the-art approach in eight maze-like environments and show that PolyFly produces faster trajectories in each scenario. We also experimentally validate our proposed approach on a real quadrotor with a suspended payload, demonstrating the practical reliability and accuracy of our method.
\end{abstract}

\vspace{-10pt}

\section*{Supplementary material}
\url{https://mrunaljsarvaiya.github.io/polyfly.github.io/}
\vspace{-5pt}

\section{Introduction}
\IEEEPARstart{M}{icro} Aerial Vehicles (MAVs), particularly quadrotors equipped with cable-suspended payload systems, have gained significant attention due to their versatility and effectiveness in diverse applications such as inspection \cite{trujillo2019}, search and rescue missions \cite{LoiannoRAL2018}, and package transportation \cite{sreenath2013geometric}. Their ability to deliver essential materials in regions inaccessible by ground robots makes them invaluable during disaster relief operations \cite{BARMPOUNAKIS2016ijtst}. Furthermore, aerial transportation systems offer considerable advantages in energy efficiency over ground vehicles for last-mile logistics, significantly reducing environmental footprints in urban delivery contexts \cite{LastMileDelivery}.

In time critical scenarios, such as rescue missions, it is crucial for aerial vehicles to optimize their flight path to reduce the mission time, especially in challenging environments like dense forests or earthquake-damaged and structurally unsafe buildings. This requires the robot planners to avoid obstacles while maneuvering through narrow gaps and around tight corners, which need non-conservative physical approximations of the system components. 

Existing methods often simplify these systems into spheres \cite{li2023autotrans} or large prisms \cite{zheng2020, tang2015mixed} to address computational demands, trading off solution quality for computational speed. In contrast, our approach directly addresses these limitations by representing each component - quadrotor, payload, and cable - as distinct polytopes. We further enhance our planner by incorporating the quadrotor’s orientation into the polytopic representation. This orientation-aware modeling enables more precise collision prediction during aggressive maneuvers in confined spaces such as the environment shown in Fig. \ref{fig:traj_9_10}. This representation provides planners with detailed geometric information, enhancing their capabilities to navigate maze-like and highly constrained environments.

Current aerial transportation planning methodologies can broadly be categorized into sampling-based and optimization-based approaches. Within optimization methods, trajectories are either parameterized as polynomials \cite{li2023autotrans,wang2024impact} or discretized and solved via Model Predictive Control (MPC) frameworks \cite{son2020, foehn2017fast, son2019, zheng2020}. Our work builds upon MPC-based global planning techniques and specifically addresses the limitations posed by existing approaches that simplify system geometry. 
Methods that employ Euclidean Signed Distance Fields (ESDFs) result in discontinuous optimization constraints \cite{schulman2014motion} while cylindrical obstacle representations limit the complexity of environments that can be modeled.
We tackle these issues by formulating obstacle avoidance explicitly through polytopic models and represent each robot component and every obstacle as distinct polytopes. As we show in this work, this representation allows us to both incorporate smooth differentiable constraints into our global planner, and generate aggressive trajectories in tightly constrained environments. 

To summarize, we present the following contributions:
\begin{itemize}
    \item A novel non-conservative representation for aerial transportation that models the quadrotor, cable, and payload as separate components. To enhance physical accuracy, we incorporate the attitude of both the robot and the obstacles into the environment representation, constructing orientation-aware polytopes that more accurately capture the system’s spatial footprint.
    \item An optimal global planning method for a robot composed of independent polytopes that produces collision-free trajectories in maze-like environments by leveraging duality theory to formulate non-linear polytopic collision constraints as smooth differentiable constraints.     
    \item Experimental validation on hardware by performing real-world experiments using a quadrotor carrying a slung payload. We illustrate the advantages of our representation and planning method by tracking trajectories in tightly constrained environments. We also demonstrate that our approach generates faster trajectories than the existing state-of-the-art method in all ten test environments. 

\end{itemize}
We release our implementation as an open-source, standalone package with minimal dependencies. 

\section{Related Works} \label{sec:related_works}
In this section, we survey the main design choices in (i) how the environment and robots are spatially represented and (ii) how the states and inputs are mathematically modeled. 

\textbf{Environment Representations.}
Obstacle-aware planners typically encode the workspace either implicitly, through Euclidean Signed Distance Fields (ESDFs), or explicitly, by representing obstacles with geometric shapes. Recent aerial transportation planners \cite{li2023autotrans, wang2024impact}, voxelize the environment and store an ESDF, turning collision checks into fast table look-ups, but introduce non-smooth, discontinuous gradients when used in optimization-based solvers.

Other works express the environment via geometric objects. \cite{foehn2017fast,son2019, son2020} employ cylindrical and spherical objects, while \cite{zheng2020, tang2015mixed, Yu2022} rely on polytopes. While ellipsoids and cylinders limit the range of scenes that can be modeled, polytopes significantly increase the complexity of environments that can be captured. In fact, works that use ESDFs also primarily consider environments composed of multiple cuboids \cite{wang2024impact}. Furthermore, in the context of trajectory planning, adopting polytopes to represent the environment yields distance constraints with good optimization properties. 
Duality theory can be used to convert non-linear polytope–to-polytope distance inequalities into smooth, differentiable constraints. \cite{pallar2025optimal, Thirugnanam2022} used polytopic collision constraints for short-horizon reactive MPC frameworks but are unable to optimize over global paths due to the short MPC horizon. \cite{zheng2020} used this approach for global planning but did not consider complex environments and cannot navigate environments through narrow gaps, eg. Fig. \ref{fig:cover_page}, due to their conservative representation that models the robot as a single cuboid. 

\begin{figure}[!t]
    \definecolor{customgreen}{HTML}{7A8607}
    \definecolor{customred}{HTML}{c17c7a}   
    \definecolor{customyellow}{HTML}{FFAB40}
    \definecolor{customblue}{HTML}{617ABA}
    \definecolor{custompink}{HTML}{B99BAA}
    \definecolor{customgray}{HTML}{7F7F7F}
    \definecolor{customlightblue}{HTML}{f1f6fd}
    \centering
    \includegraphics[width=0.65\columnwidth,
                   trim=110 70 350 40, clip]{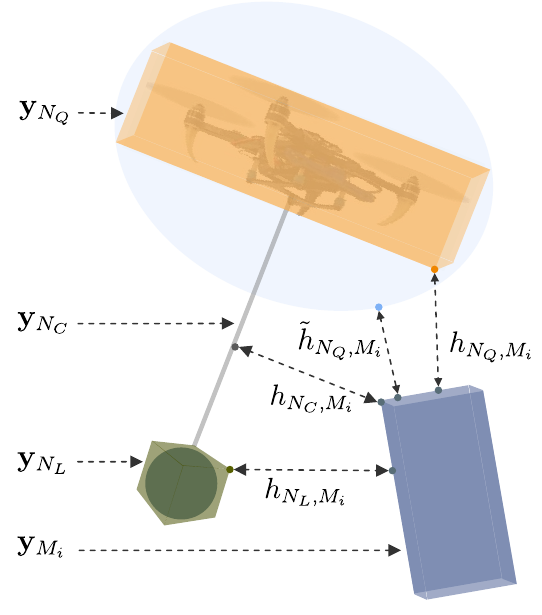}
    \caption{Different polytopes used in PolyFly: \textbf{\textcolor{customyellow}{Quadrotor}} $N_Q$, 
    \textbf{\textcolor{customgreen}{Payload}} $N_L$,
   \textbf{\textcolor{customgray}{Cable}} $N_C$, and
   \textbf{\textcolor{customblue}{Obstacles}} $M_i$. The quadrotor polytope enables accurate distance computations ${h}_{N_Q,M_i}$, whereas most methods use a sphere around the quadrotor resulting in conservative distance computations $\tilde{h}_{N_Q,M_i}$.}
    \label{fig:poly}
    \vspace{-10pt}
\end{figure}
\textbf{Spatial Robot Representation.}
There are two primary design decisions with respect to modeling the robot. First, how should the physical geometry of the robot be modeled? Existing methods typically use multiple spheres or ellipses to represent the quadrotor and its payload \cite{li2023autotrans, wang2024impact, foehn2017fast,son2019, son2020} or a single prism or polytope \cite{zheng2020, tang2015mixed}. These representations over-approximate at least one component (quadrotor, payload, cable or environment obstacles), therefore limiting their applicability in highly constrained or cluttered environments. Spheres over-approximate the quadrotor, cable, and obstacles while ellipses over-approximate the quadrotors and obstacles used in this work. For example, ellipsoidal representations fail in the ceiling-wall environments in Fig. \ref{fig:ceiling_examples}, while single polytope representations \cite{zheng2020} fails in the ceiling-floor and narrow gap environments in Fig. \ref{fig:traj_9_10} and \ref{fig:ceiling_examples}.  In contrast, our method addresses these limitations by leveraging orientation-aware polytopes for each robot component - quadrotor, cable, and payload. Combined with polytopic representation of the environments, we enable non-conservative distance evaluation, as shown in Fig. \ref{fig:poly}, facilitating aerial transportation in maze-like environments and narrow passages that are slightly larger than the width of the cable.  

\textbf{State and Input Modeling.}
The second design decision is how states and inputs are mathematically represented. A popular strategy is to express the states and inputs as polynomials to facilitate trajectory optimization \cite{li2023autotrans, wang2024impact}. This is often paired with MINCO \cite{wang2022minco} to reformulate the problem into an unconstrained optimization to boost computational efficiency. While this approach can be used to achieve real-time capabilities, our experiments in Section \ref{subsec:q1} indicate that it may compromise both solution quality and convergence. In contrast, similar to \cite{foehn2017fast, zheng2020, son2019, son2020}, PolyFly operates directly in the native state space. Unlike polynomial parameterizations, the native state space approach places no restrictions on the shape of the state or control trajectories, making them inherently more expressive. 

The method presented in \cite{li2023autotrans, wang2024impact}, serves as our primary baseline, employs an ESDF to represent the environment, approximates the quadrotor and payload with spheres, and parameterizes the states and inputs via polynomials. \cite{wang2024impact} reports an ablation study comparing the computational times and success rates of their method with IPOPT-MP, a state-based global planning approach. However, they omit the comparison of the optimized trajectory times of their method with IPOPT-MP, and do not address navigation close to obstacles in maze-like environments. We demonstrate that by using the native state-based modeling and a non-conservative spatial workspace representation via polytopes, PolyFly outperforms this method in all test environments.
\section{System Dynamics} \label{sec:system_dynamics}
\begin{table}[!t]
\caption {Notation table\label{tab:notation}} 
\centering
\begin{tabularx}{0.48\textwidth}{>{\hsize=0.65\hsize}X >{\hsize=1.42\hsize}X}
\hline 
$\worldf$, $\robotf{}$ & inertial, and robot frame\\
$\loadmass,m_Q \in \realnum{}$ & mass of payload and robot\\
$\loadpos{},\robotpos{} \in \realnum{3}$ & position of payload and robot in $\worldf$\\
$\loadvel,\loadacc, \loadjerk \in \realnum{3}$ & payload velocity, acceleration, jerk in $\worldf$\\
$\robotvel,\robotacc \in \realnum{3}$ & robot velocity, acceleration in $\worldf$\\
$\robotrot{} \in\SOthree$& robot orientation with respect to $\worldf$ \\
$\robotangvel{}\in\realnum{3}$& robot angular velocity in $\robotf{}$\\
$f_{}\in\realnum{}$, $\matM_{}\in\realnum{3}$&collective thrust and moment on robot in $\robotf{}$\\
$J \in \mathbb{R}^{3 \times 3}$ & moment of inertia of robot in $\robotf{}$\\
$\cablevec{}\in S^2$&unit vector from robot to payload in $\worldf$\\
$f_i, l_{}, g\in\realnum{}$ & $i^{\text{th}}$ motor thrust, cable length, gravity\\
$N_O \in \mathbb{Z}$ & total number of obstacles\\
$M_i$, $N_j$& $i^{th}$ obstacle and $j^{th}$ robot polytope, \\
\hline
\end{tabularx}
    \vspace{-20pt}
\end{table}
In this section, we summarize the non-linear dynamics of a quadrotor carrying a suspended payload. We follow the derivation presented in \cite{sreenath2013geometric} and assume that the cable remains taut. Applying the Lagrange–d’Alembert principle, we obtain
\begingroup              
\allowdisplaybreaks      
\begin{align}
\frac{d\loadpos}{dt} = \loadvel, ~\frac{d\robotpos{}}{dt} &= \robotvel{}, ~\dot{\mathbf{R}}_{Q} = \robotrot{} \hat{\mathbf{\Omega}},\label{eq:single-kinematics} \\
\prths{m+\loadmass}\prths{\loadacc + \vecg} &= \prths{\cablevec{}\cdot f\robotrot{}\axis{3}{}-m l\prths{\cabledotvec{}\cdot\cabledotvec{}}}\cablevec{},\label{eq:single-load-lagrange-eom} \\
m l\prths{\cableddotvec{}+\prths{\cabledotvec{}\cdot\cabledotvec{}}\cablevec{}} & = \cablevec{}\times\prths{\cablevec{}\times f\robotrot{}\axis{3}{}},\label{eq:single-quad-lagrange-eom}\\
\matM &= \inertia\robotangacc{} + \robotangvel{}\times\inertia\robotangvel{},\label{eq:single-robot-rotation-dyn}\\
\begin{bmatrix}
  f,
  \mathbf{M}
\end{bmatrix}^\top 
&= \mathbf{D}
\begin{bmatrix}
  f_{1}, f_{2}, f_{3}, f_{4}
\end{bmatrix}^\top \label{eq:thrust_map}.
\end{align}
\endgroup
where $\vecg = g \axis{3}{}$, $g = 9.81$ $m/s^{2}$, $\axis{3}{} = \left[0~0~1\right]^{\top}$, and $\hat{\robotangvel{}}$ is the skew-symmetric matrix of the quadrotor angular velocity $\robotangvel{}$. Table \ref{tab:notation} defines the remaining variables. Qualitatively, eq.~\eqref{eq:single-load-lagrange-eom} describes the relationship between the payload acceleration and collective motor thrust, while eq.~\eqref{eq:single-quad-lagrange-eom} shows that cable's angular acceleration depends on the collective thrust and the quadrotor's orientation. The matrix $\mathbf{D}$ provides a linear mapping between collective thrust-moments and individual motor thrusts. It is defined in terms of the robot's physical properties and propellers aerodynamics characteristics. Readers can refer to \cite{lee2010} for additional information.

By leveraging the differential flatness property of the system \cite{sreenath2013geometric}, we can model the state dynamics described in eqs.~\eqref{eq:single-kinematics}--\eqref{eq:thrust_map} in terms of the flat outputs $\{\loadpos{}, \psi\}$, where $\psi$ is the quadrotor's yaw angle. In our method, we set $\psi = 0$ and assume negligible cable angular accelerations. This allows us to express the system through the state $\vecx = \begin{bmatrix}\loadpos{}^\top, \loadvel{}^\top, \loadacc{}^\top \end{bmatrix}^\top \in \realnum{9}$ and input $\vecu = \begin{bmatrix} \loadjerk \end{bmatrix} \in \realnum{3}$.

The induced system dynamics are therefore modeled as a third-order system. Given $\vecx$, $\vecu$, the robot's physical parameters, and assuming that cable angular accelerations are zero at the states optimized by the planner, we can use the non-linear differential flatness mapping \cite{sreenath2013geometric} to calculate the robot position $\robotpos{}$, robot velocity $\robotvel{}$, and orientation $\robotrot{}$. This allows our trajectory generator to plan in payload space and add constraints to the quadrotor's states. Since our planner operates in a discrete space, we convert the continuous time representation above to its discrete form. 
We obtain the Runge-Kutta approximation $\vecx(k+1) = F(\vecx(k), \vecu(k), \Delta t_k)$, where $F$ is the discretized dynamics function and $\Delta t_k$ the time step duration at stage $k$.
\section{Planning} \label{sec:planning}
\subsection{Preliminaries} \label{subsec:prelim}
We develop a global planner that computes a collision-free trajectory $\boldsymbol{\tau}$ for a quadrotor with a suspended payload where
\begin{equation}
\boldsymbol{\tau} ={} \begin{bmatrix}\boldsymbol{\tau}_0,\dots,\boldsymbol{\tau}_N\end{bmatrix},
\boldsymbol{\tau}_k ={} \begin{bmatrix}\loadpos(k),\,\loadvel(k),\,\loadacc(k)\end{bmatrix}^{\top},
\end{equation}
where $k \in (0, N)$. The robot starts at state $\vecx_O$ and must navigate to its goal state $\vecx_N$. The environment is described by the set $\mathbf{E}$, which contains $N_O$ polytopic obstacles $M_i$ for $i \in (0, N_O)$. The aerial system is modeled  using three separate polytopes, $N_j$ for $\; j \in \{Q, C, L\}$ that represent the quadrotor, cable, and load polytopes respectively, as  shown in Fig. \ref{fig:poly}. 

\subsection{Polytopic Approximation}
Let us consider two polytopes $\mathbf{M_i}$ and $\mathbf{N_j}$, as defined in Table \ref{tab:notation}. Let $\ybf{M_i}$ be any point within the polytope $\mathbf{M_i}$ and $\ybf{N_j}$ be any point within the polytope $\mathbf{N_j}$. $\mathbf{A_{M_i}}, \mathbf{B_{M_i}}, \mathbf{A_{N_j}}$ and $\mathbf{B_{N_j}}$ are constants that define their geometric shape. These polytopes can be represented by 
\begin{equation}
\label{eq:poly_M_rep}
\mathbf{A}_{M_i} \ybf{M_i}\le\mathbf{B}_{M_i},~
\mathbf{A}_{N_j}\ybf{N_j}\le\mathbf{B}_{N_j}.
\end{equation}
While not strictly necessary, we simplify computation by representing one polytope's points in the other polytope's frame. We represent the obstacle points $\ybf{M_i}$ using the relative position vector $\ybf{M_i} ^ {N_j}$, which gives
\begin{equation}
    \ybf{M_i} = \rot{N_j} \ybf{M_i} ^ {N_j} + \mathbf{O}_{N_j},
\end{equation}
where $\rot{N_j}$ is ${N_j}$'s rotation matrix, and $\mathbf{O}_{N_j}$ is its origin. Substituting this into the polytopic form expressed by eq.~(\ref{eq:poly_M_rep}), we can represent polytope $M_i$ as
\begin{align}
    (\mathbf{A}_{M_i} \rot{N_j}) \ybf{M_i}^{N_j} &\leq \mathbf{B_{M_i}} - \mathbf{A}_{M_i} \mathbf{O}_{N_j}.
\end{align}
Since $\rot{N_j}$ and $\mathbf{O}_{N_j}$ are functions of the chosen state vector $\vecx$, $M_i$ is represented by 
\begin{align}
    \mathbf{A}_{M_i}^{N_j}(\vecx) \ybf{M_i}^{N_j} &\leq \mathbf{B}_{M_i}^{N_j} (\vecx), \label{eq:a_b_rep} \\
    \mathbf{A}_{M_i}^{N_j}(\vecx) = \mathbf{A}_{M_i} \rot{N_j}(\vecx)&, \mathbf{B}_{M_i}^{N_j} (\vecx) = \mathbf{B_{M_i}} - \mathbf{A}_{M_i} \mathbf{O}_{N_j}(\vecx) \notag.
\end{align}
\subsection{Distance Metric Between Polytopes \label{subsec:distance_poly}}
Our optimization problem seeks to lower bound $h_{M_i,N_j}$, the distance between polytopes $M_i$ and $N_j$, with a positive margin $\beta$. Specifically, 
\begin{equation}
h_{M_i,N_j}(\vecx) \geq \beta, ~h_{M_i,N_j}(\vecx) := d(M_i, N_j),\label{eq:dist_payload_obs_ineq} 
\end{equation}
where $d$ is the distance function described in \cite[Eq.~(8a)]{zhang2020optimization}. The constraint defined by eq. (\ref{eq:dist_payload_obs_ineq}) is non-differentiable and difficult to integrate into optimization problems when the obstacles and robots are represented as convex polytopes \cite{zhang2020optimization}. To handle this complexity, we follow the approach in \cite[Proposition~1]{zhang2020optimization}. and reformulate eq.~(\ref{eq:dist_payload_obs_ineq}) using duality theory into the following set of smooth non-linear differentiable constraints. The equivalent dual problem is 
\begin{align}
g_{M_i,N_j}(\vecx) &= \max_{\mathbf{\lambda}_{N_j}, \mathbf{\lambda}_{M_i}} -\mathbf{\lambda}_{N_j}^\top \mathbf{B}_{N_j} - \mathbf{\lambda}_{M_i}^\top \mathbf{B}_{M_i}^{N_j}(\vecx) \label{eq:dual}\\
\text{s.t.} \quad &\mathbf{\lambda}_{N_j}^\top \mathbf{A}_{N_j} + \mathbf{\lambda}_{M_i}^\top \mathbf{A}_{M_i}^{N_j}(\vecx) = 0 \notag\\
&\mathbf{\lambda}_{M_i} \geq 0, \quad \mathbf{\lambda}_{N_j} \geq 0, \quad \left\| \mathbf{\lambda}_{M_i} * \mathbf{A}_{M_i}  \right\|_2 \leq 1, \notag
\end{align}
where $\mathbf{\lambda}_{N_j} \in \realnum{n_j}$ and $ \mathbf{\lambda}_{M_i} \in \realnum{m_i}$ are the dual variables associated with the obstacle avoidance problem. For our setup, $m_i=n_j=6$. Results in \cite{zhang2020optimization,boyd2004convex} show that Strong Duality holds for the dual problem defined in eq. (\ref{eq:dual}), which gives us the lower bound of the primal cost function $h_{M_i,N_j}(\vecx)$
\begin{equation}
\bar{g}_{M,N}(\vecx, \mathbf{\lambda}_{M_i}, \mathbf{\lambda}_{N_j}) := -\mathbf{\lambda}_{N_j}^\top \mathbf{B}_{N_j} - \mathbf{\lambda}_{M_i}^\top \mathbf{B}_{M_i}^{N_j}(\vecx) \leq h_{M_i,N_j}(\vecx).
\label{eq:dual_cost_bound}
\end{equation}

Therefore, to enforce a minimum distance of $\beta$ between $M_i$ and $N_j$, we constrain the lower bound $\bar{g}_{M,N}\geq \beta$, giving us a smooth non-conservative reformulation of eq.~(\ref{eq:dist_payload_obs_ineq})
\begin{equation}
\begin{aligned}
&-\mathbf{\lambda}_{N_j}^\top \mathbf{B}_{N_j} - \mathbf{\lambda}_{M_i}^\top \mathbf{B}_{M_i}^{N_j}(\vecx) \geq \beta, \label{eq:final_dual} \\
\quad &\mathbf{\lambda}_{N_j}^\top \mathbf{A}_{N_j} + \mathbf{\lambda}_{M_i}^\top \mathbf{A}_{M_i}^{N_j}(\vecx) = 0, \\
&\mathbf{\lambda}_{M_i} \geq 0, \quad \mathbf{\lambda}_{N_j} \geq 0, \quad \left\| \mathbf{\lambda}_{M_i} * \mathbf{A}_{M_i}  \right\|_2 \leq 1. \\
\end{aligned}
\end{equation}

We add the set of constraints defined in eq. (\ref{eq:final_dual}) for collisions between each $(M_i, N_j)$ pair, where $i \in (0, N_O)$. For notational convenience, we combine the dual variables into a vector $\mathbf{\Lambda} \in \realnum{Q}$, where $Q = N \times N_O \times N_R \times (m_i + n_j)$, and $N_R=3$ is the number of robot components.

\subsection{Optimization Problem Formulation}
Our planner solves the following optimization problem
\begin{align}
\min_{\vecx(k), \vecu(k), \Delta t_k, \mathbf{\Lambda}}& \sum_{k=0}^{N-1} L(\vecx(k), \vecu(k), \Delta t_k) \label{eq:nmpc_ineq_constraint_x}\\
\text{subject to \, \, \,} \vecx(k+1) &= F(\vecx(k), \vecu(k), \Delta t_k, \notag\\
\vecx(0) &= \vecx_0, ~ \vecx(N) = \vecx_N, \notag\\
H_l(\vecx(k),\vecu(k),\mathbf{\Lambda}) &\leq 0, ~\Delta t_{min} \leq \Delta t_k \leq \Delta t_{max}, \notag
\end{align}

where $ \forall k \in (0, N-1)$, $\forall l \in (0,N_h)$, $N$ is the horizon length, $\Delta t_k$ is the time duration for stage $k$, and $x_0$ and $x_N$ are the initial and terminal states respectively. $\Delta t_{min}$ and $\Delta t_{max}$ are the min. and max. time durations set by the user and are discussed in Section \ref{subsubsection:time_min}. $H_l$ comprises the $N_h$ inequality constraints from the collision-avoidance formulation in eq.~\eqref{eq:final_dual} together with the state and input bounds specified in Section~\ref{subsec:state_input_constr}. Finally, $L = L_t + L_u + L_g + L_{td}$ is the loss function defined in the subsequent subsections. 
 
We now describe the optimization constraints, costs and initialization methods. The variables $\Delta t_{min}$, $\Delta t_{max}$, $\alpha_{to}$, $\alpha_{u}$, $\alpha_{g}$, $\alpha_{td}$ and $N$, are user defined constants whose values are defined in Section \ref{sec:experiments}.

\subsubsection{Time Minimization}
\label{subsubsection:time_min}
We incentivize the planner to minimize the total trajectory time by adding a large cost to each time step duration. Specifically,
\begin{equation}
    L_{t} = T \frac{\alpha_{to}}{N}, T = \sum_{k=0}^{N-1} \Delta t_k,
    \Delta t_{min} \leq \Delta t_k \leq \Delta t_{max},
    \label{eq:minimize_sum}
\end{equation}
where $\Delta t_{min}, \Delta t_{max}, \alpha_{to} > 0$.

\subsubsection{Input Smoothness Regularization}
We add a small cost that penalizes large changes between consecutive inputs, thereby regularizing their rate of change and improving the smoothness of the resulting robot states.
\begin{equation}
    L_{u} = \frac{\alpha_{u}}{N} \sum_{k=1}^{N} \| \mathbf{\vecu}_i - \mathbf{\vecu}_{i-1} \| ^2, \quad \alpha_{u} > 0.
\end{equation}

\subsubsection{Regularizing Proximity to the Initial Guess} \label{subsubsection:reg_proxy}
Even though we use a simple A* initialization, we expect the optimal trajectory's positions to lie in the vicinity of the initialized positions. We achieve this by adding a small penalty to deviations of the optimized positions from the initial guess. The study supporting the benefits of this term is discussed in Section \ref{subsec:prox_guess}.
\begin{equation}
    L_{g} = \frac{\alpha_{g}}{N-1} \sum_{k=1}^{N-1} \| \vecx(k) - \vecx_{g}(k) \|  ^2 ,  
\end{equation}
where $\vecx_g(k)$ is the initial guess at stage index $k$ and $\alpha_{g} > 0$.

\subsubsection{Regularizing Subsequent Time Step Durations}
We add a small cost to the difference between subsequent time step durations to help our solver converge. Specifically, we add
\begin{equation}
    L_{td} = \frac{\alpha_{td}}{N} \sum_{k=0}^{N-1} \| \Delta t_{k+1} - \Delta t_k \| ^2, \quad \alpha_{td} > 0.
\end{equation}

\subsubsection{Solver Initialization}
\begin{figure}[!t]          
  \centering
  \begin{minipage}[b]{.49\columnwidth}
    \includegraphics[width=\linewidth, trim=0 200 60 100, clip]{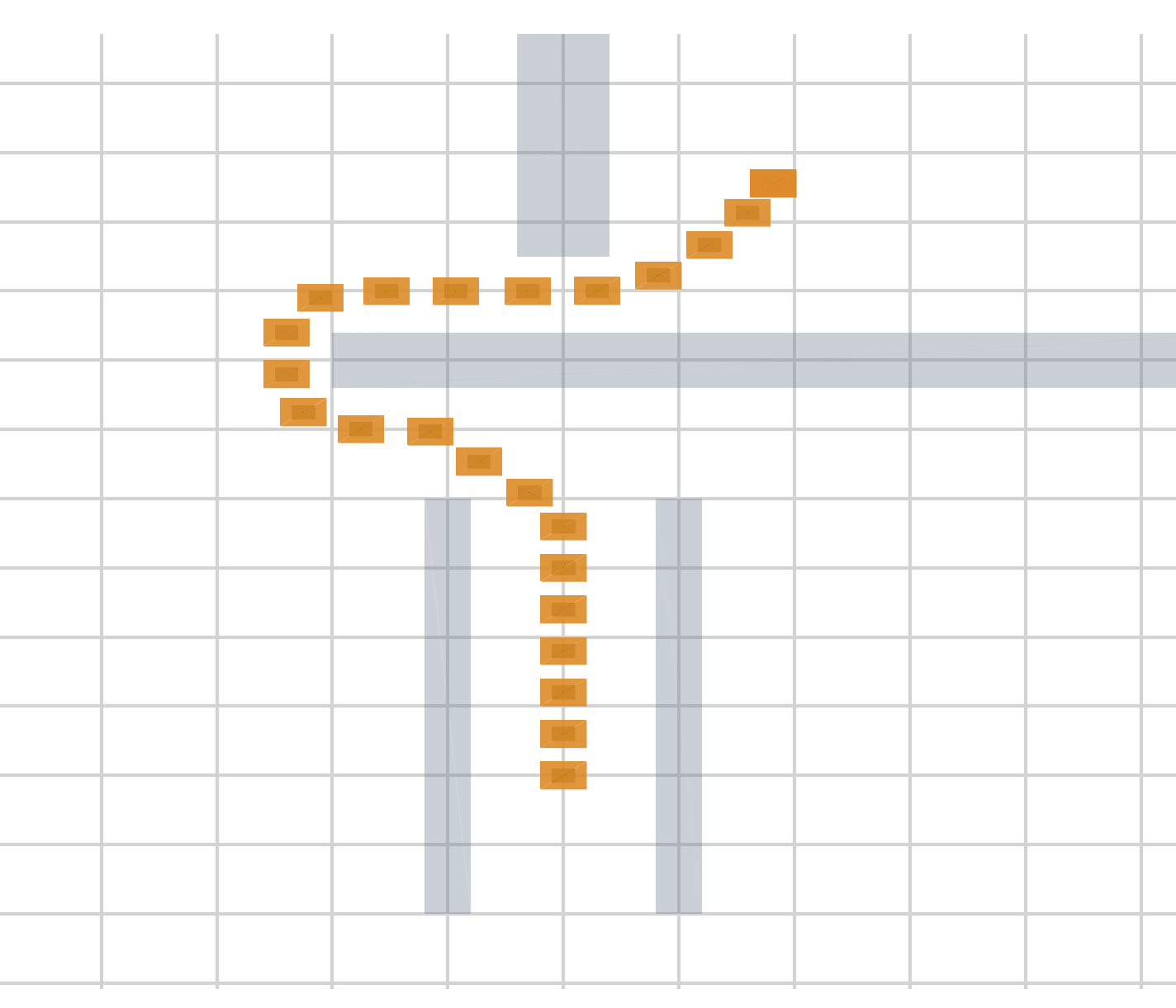}
  \end{minipage}\hfill%
  \begin{minipage}[b]{.49\columnwidth}
    \includegraphics[width=\linewidth, trim=0 210 60 100, clip]{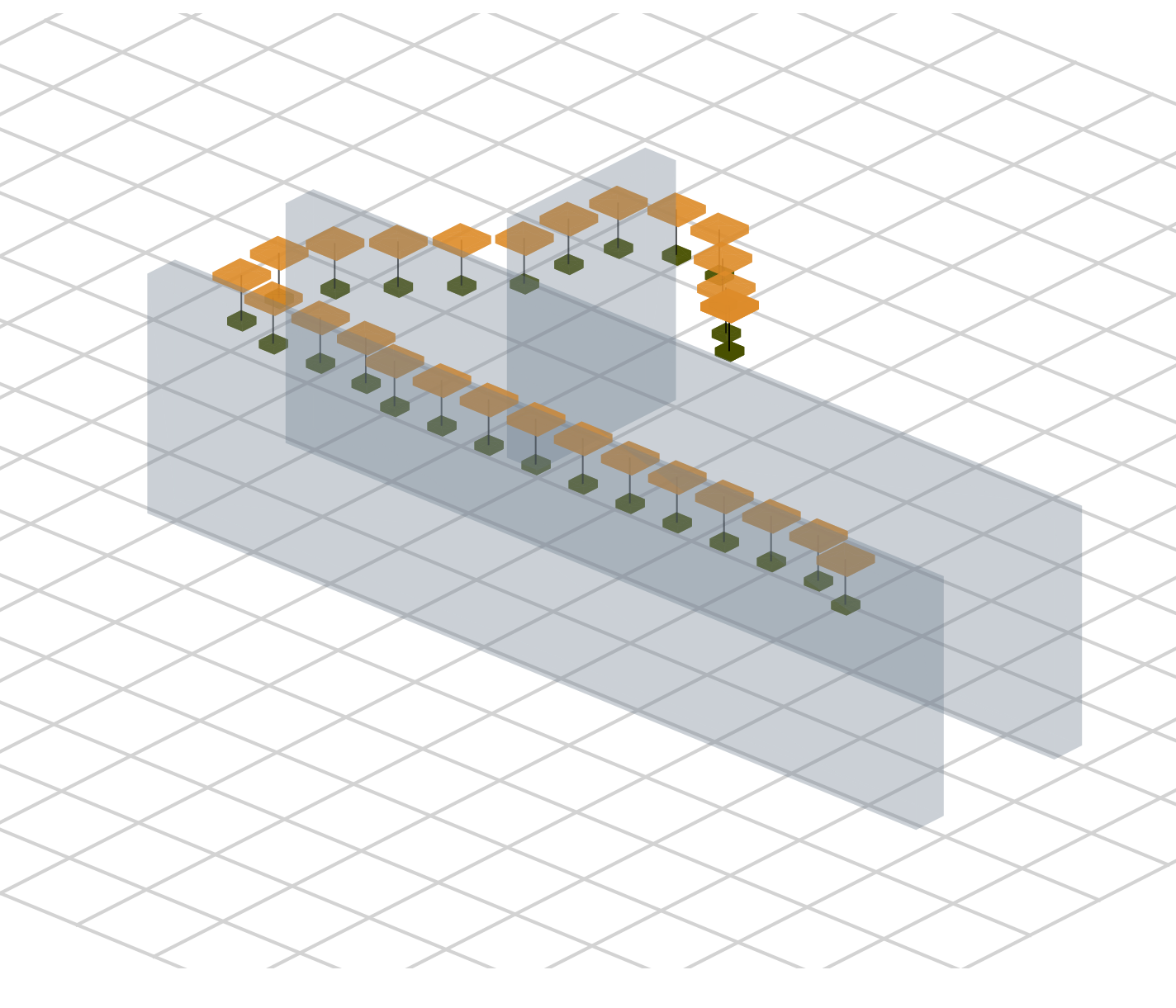}
  \end{minipage}
  \caption{A* trajectories for Env. 6 (left) and Env. 8 (right)}
  \label{fig:traj_init_astar}
      \vspace{-15pt}
\end{figure}
Since the optimization problem combines non-linear dynamics with non-convex collision constraints, the resulting solution is sensitive to its initial guess. Whereas many prior planners \cite{wang2024impact, li2023autotrans} adopt a kino-dynamic initialization strategy, our approach begins from a simple A* path that ignores the non-linear dynamics and assumes the cable is always aligned with the global $z$-axis, i.e., $\loadacc = 0$. Fig. \ref{fig:traj_init_astar} visualizes this initialization for two representative environments. 
We then use a lightweight initialization method for the load velocities using the equation 
\begin{equation}
\loadvel{_{,k}} = \frac{\loadpos{_{,k+1}} - \loadpos{_{,k}}}{\Delta t_{max} - \Delta t_{min}},
\label{eq:vel_init}
\end{equation}
where $ 0 < \Delta t_{min} < \Delta t_{max}$. Even with this relatively simple initialization method, our planner produces high quality trajectories that are consistently better than the baseline.

\subsubsection{Collision Constraints}
We account for the collision between the quadrotor, cable, and payload polytopes against each obstacle. For each time stage $k$ and each polytope pair, we add the constraints derived in Section \ref{subsec:distance_poly}. As an example, to handle collisions between the payload polytope $N_L$ and an obstacle $M_i$, we add the following constraints for all stages $k$ and obstacles $i$
\begin{align}
\text{$\forall k,~\forall i$ \quad} &-\mathbf{\lambda}_{N_L, k}^\top \mathbf{B}_{N_L} - \mathbf{\lambda}_{{M_i}, k}^\top \mathbf{B}_{M_i} ^ {N_L}(\vecx_k) \geq \beta,  \label{eq:dual2}\\
&\mathbf{\lambda}_{N_L, k}^\top \mathbf{A}_{N_L} + \mathbf{\lambda}_{M_i, k}^\top \mathbf{A}_{M_i}^{N_L}(\vecx_k) = 0, \notag\\
&\mathbf{\lambda}_{M_i, k} \geq 0, \quad \mathbf{\lambda}_{N_L, k} \geq 0, \quad \left\| \mathbf{\lambda}_{M_i, k} * \mathbf{A}_{M_i} \right\|_2 \leq 1. \notag
\end{align}
\subsubsection{State and Input Constraints} \label{subsec:state_input_constr}
The system’s initial and final positions are specified based on the environment and are enforced via equality constraints. We also constrain the start and end velocities, accelerations and jerks to zero to account for hardware limitations. Specifically, 
\begin{equation}
\begin{aligned}
    \vecx(0)  = \vecx_0, \ \vecx(N)  = \vecx_N, ~ \vecu(0) = \vecu(N) &= 0,\\
    \loadvel(0) = \loadvel(N) = 0,~ \loadacc(0) = \loadacc(N) &= 0,\\
    \robotvel(0) = \robotvel(N) = 0,~ \robotacc(0) = \robotacc(N) &= 0.
\end{aligned}
\end{equation}
Finally, we bound the state and inputs based on the robot's limitations and the environment dimensions.
\begin{align}
    \vecx_{l} \leq \vecx \leq \vecx_{h}, ~
    \vecu_{l} \leq \vecu \leq \vecu_{h},
\end{align}
where $\vecx_{l}, \vecu_{l}$ and $\vecx_{h}, \vecu_{h}$ are the lower and upper bounds. 
\subsubsection{Incorporating the Quadrotor's Rotation into the Polytopic Constraints}
To improve the accuracy of the polytopic representation, we rotate the quadrotor polytope by the quadrotor's rotation. As discussed in Section \ref{sec:system_dynamics}, we can derive the robot's rotation $\robotrot$ using $\vecx$ and $\vecu$, and use it to compute the $\mathbf{A}_{M_i} ^ {N_Q}$ matrix in eq. (\ref{eq:a_b_rep}).

\section{Experimental Results}  \label{sec:experiments}

\definecolor{customgreen}{HTML}{7A8607}
\definecolor{customred}{HTML}{c17c7a}   
\definecolor{customyellow}{HTML}{FFAB40}
\definecolor{customlightblue}{HTML}{9ba4bd}
\definecolor{customdarkgray}{HTML}{4a4a4a}
\begin{figure*}[!t]          
  \centering
  \begin{minipage}[b]{.24\textwidth}
    \includegraphics[width=\linewidth]{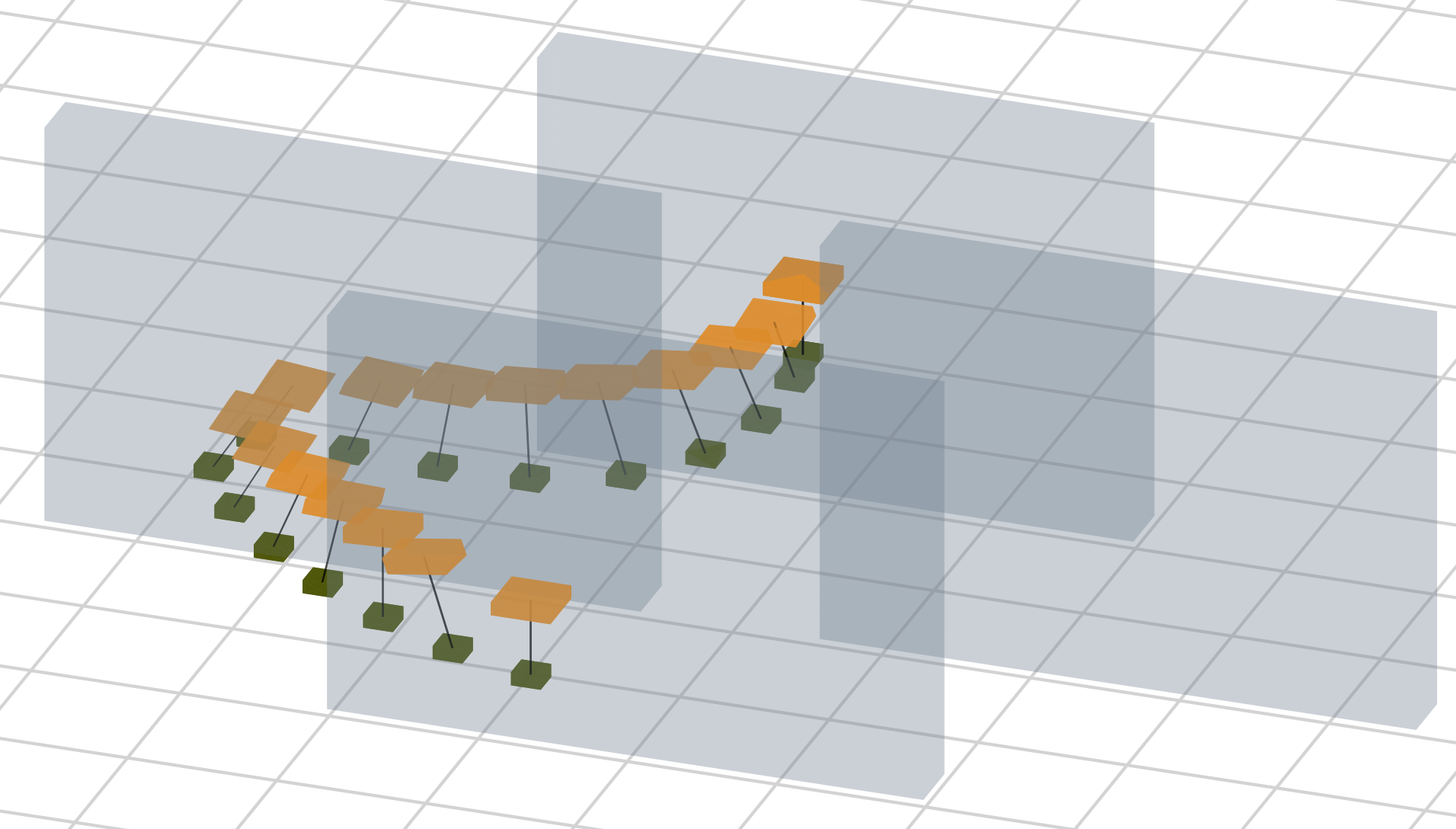}
  \end{minipage}\hfill%
  \begin{minipage}[b]{.24\textwidth}
    \includegraphics[width=\linewidth]{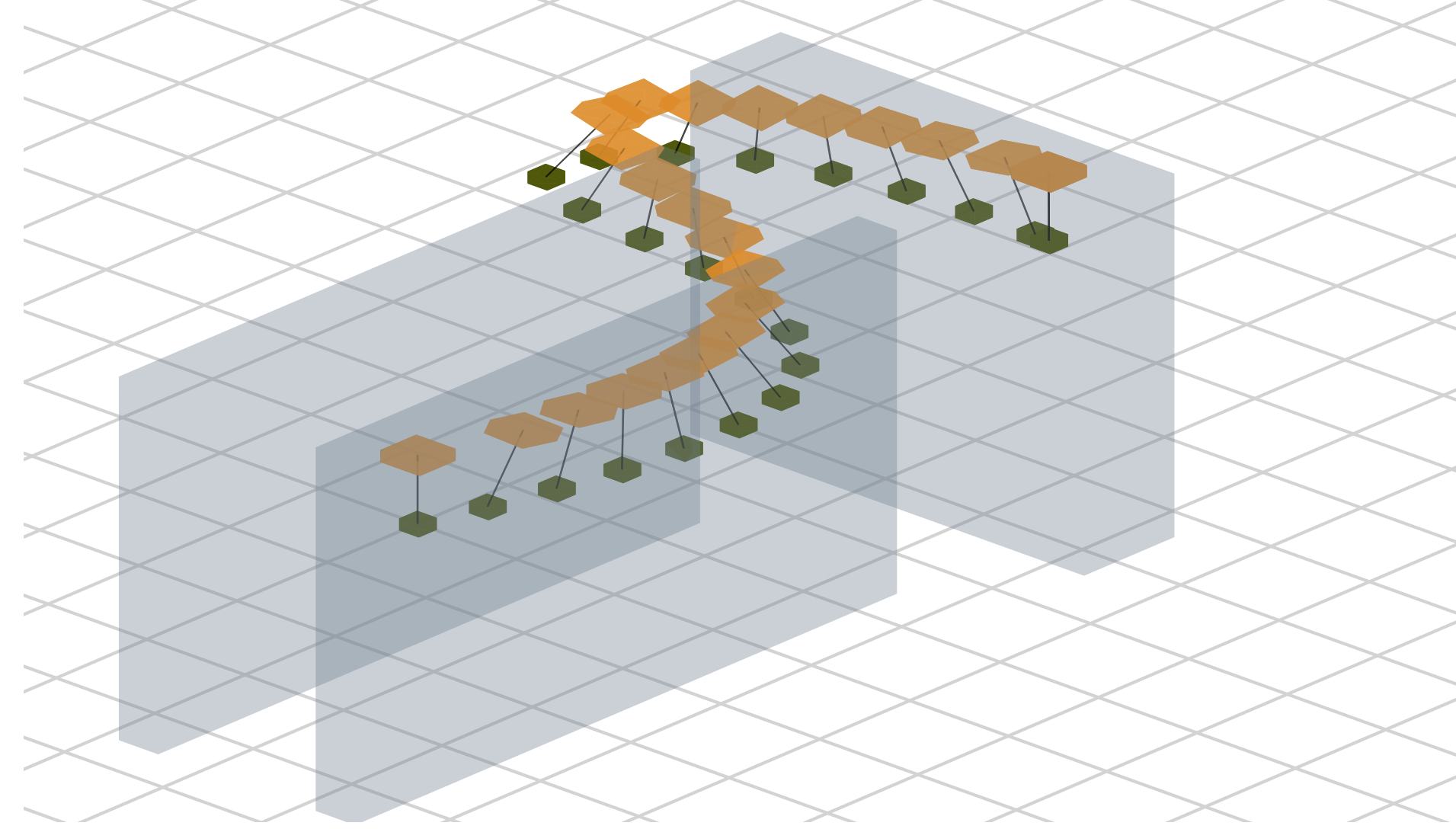}
  \end{minipage}\hfill%
  \begin{minipage}[b]{.24\textwidth}
    \includegraphics[width=\linewidth]{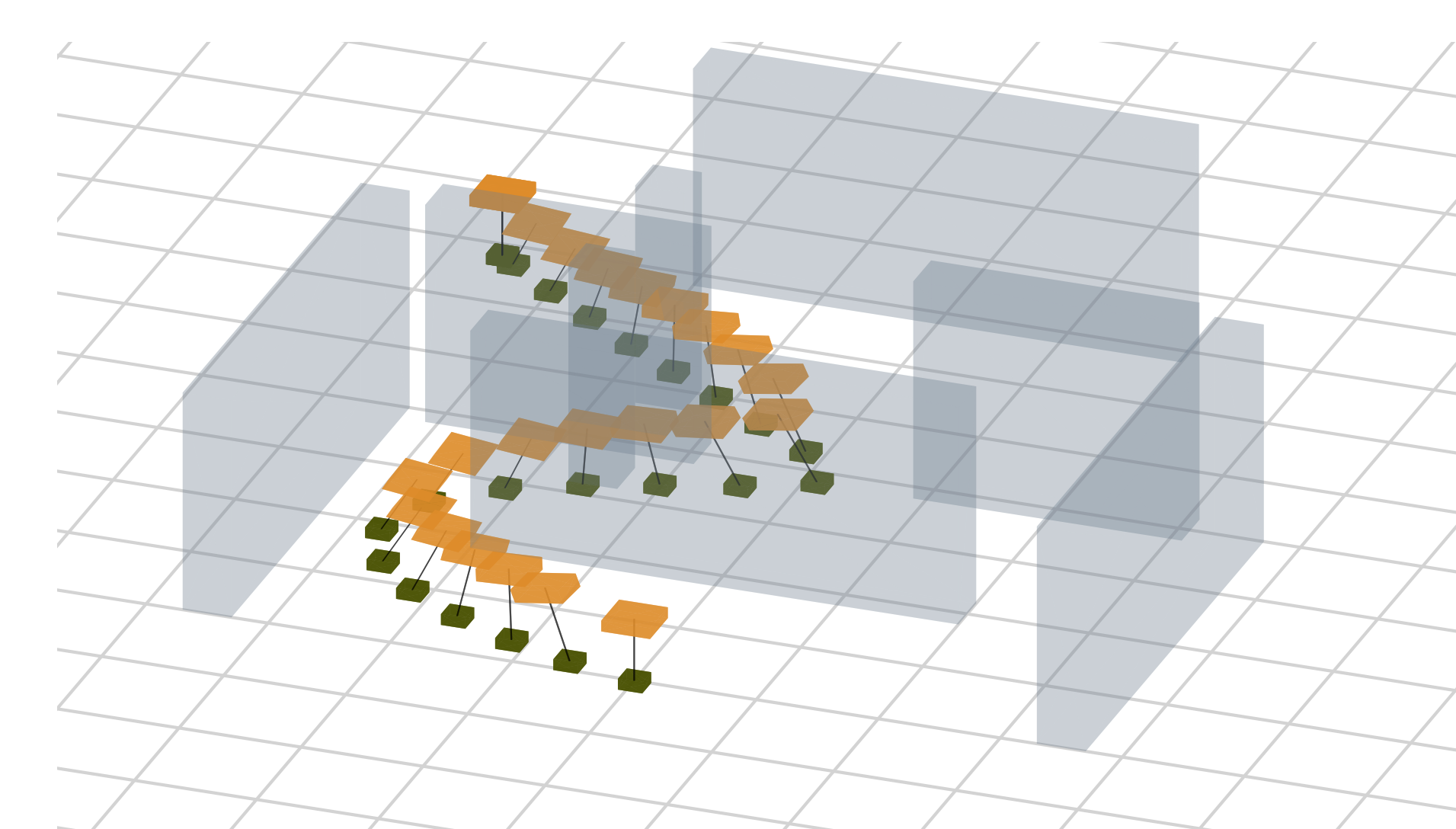}
  \end{minipage}\hfill%
  \begin{minipage}[b]{.24\textwidth}
    \includegraphics[width=\linewidth]{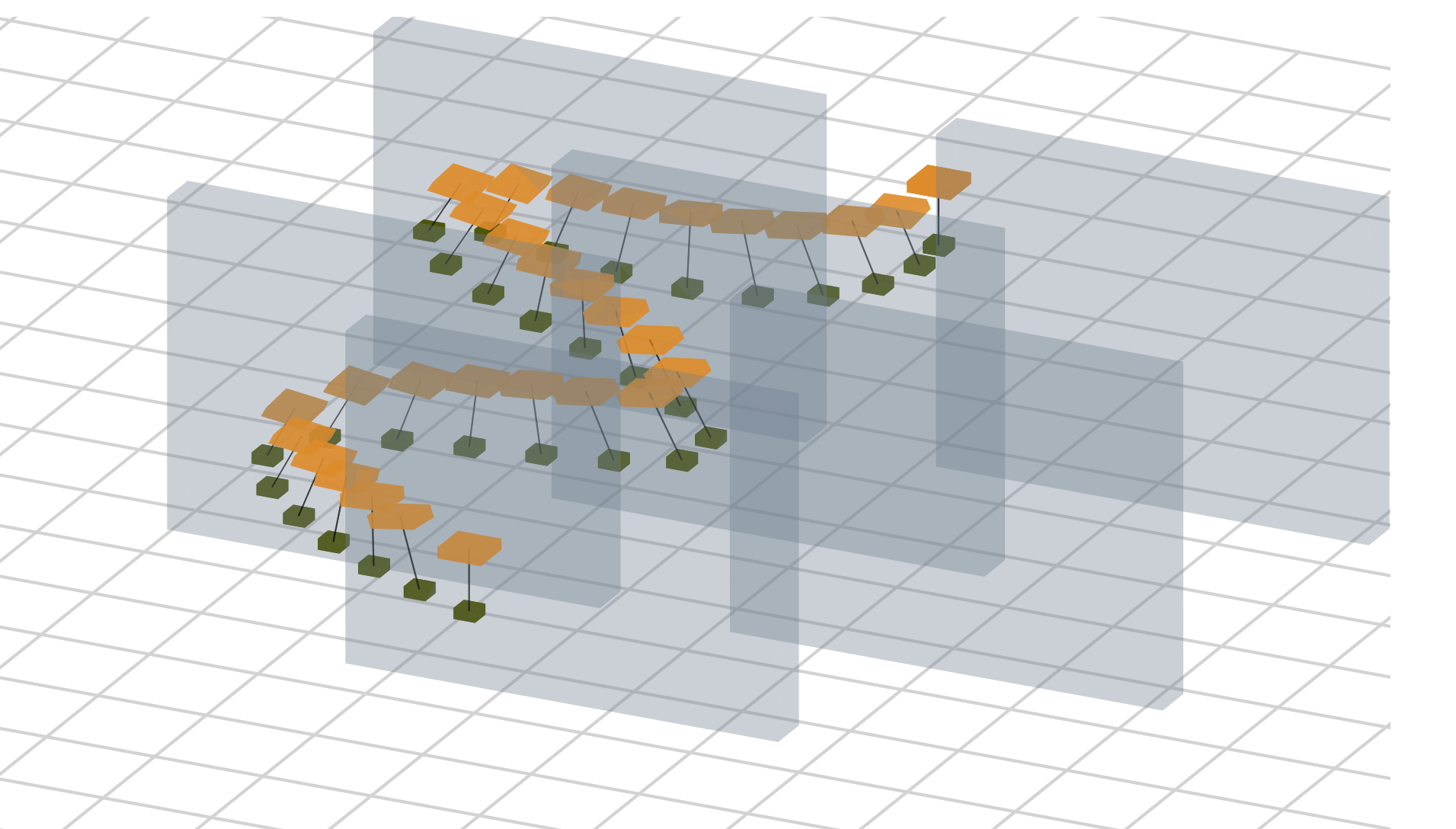}
  \end{minipage}

  \vspace{3pt}
  \begin{minipage}[b]{.24\textwidth}
    \includegraphics[width=\linewidth]{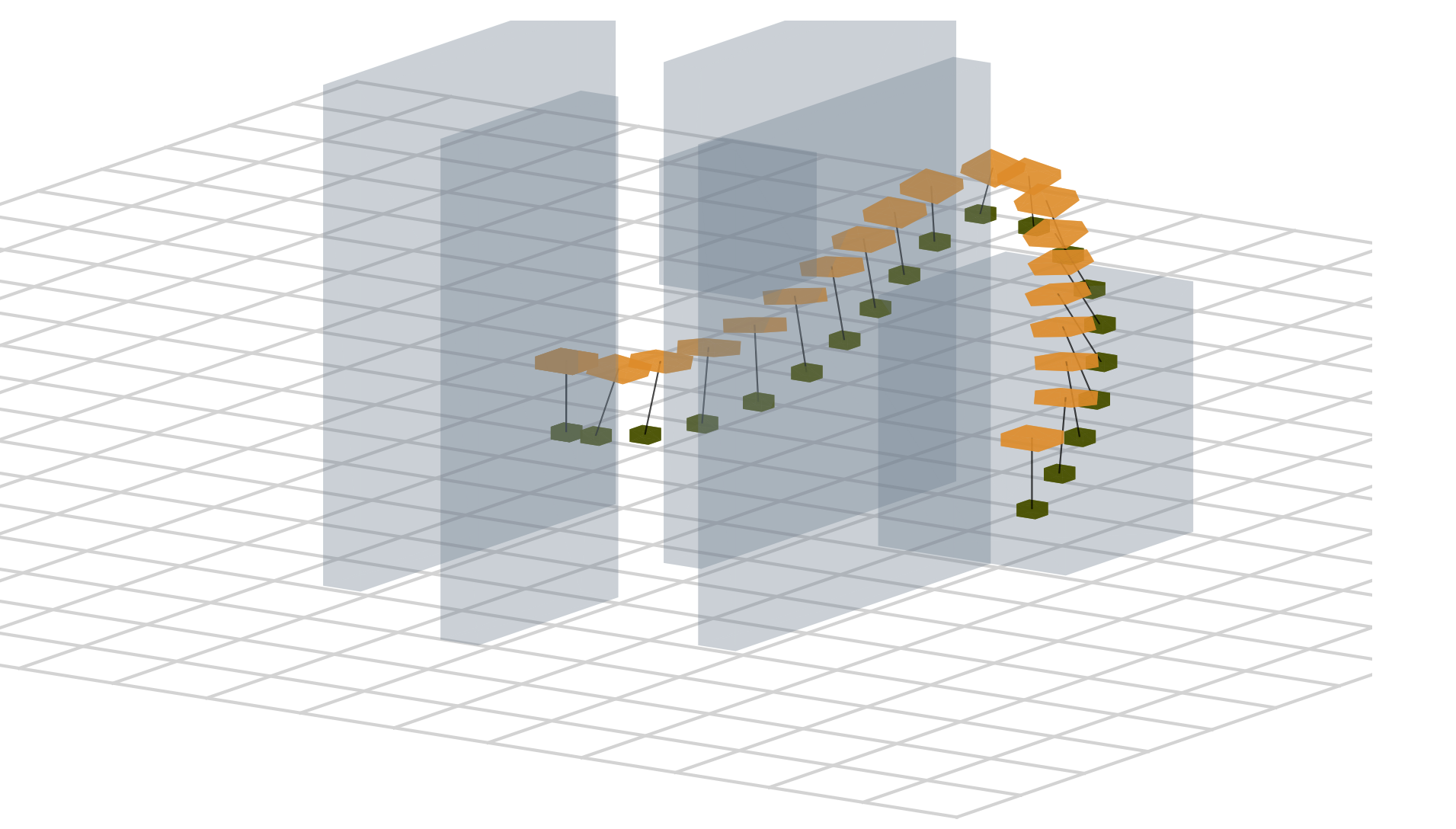}
  \end{minipage}\hfill%
  \begin{minipage}[b]{.24\textwidth}
    \includegraphics[width=\linewidth]{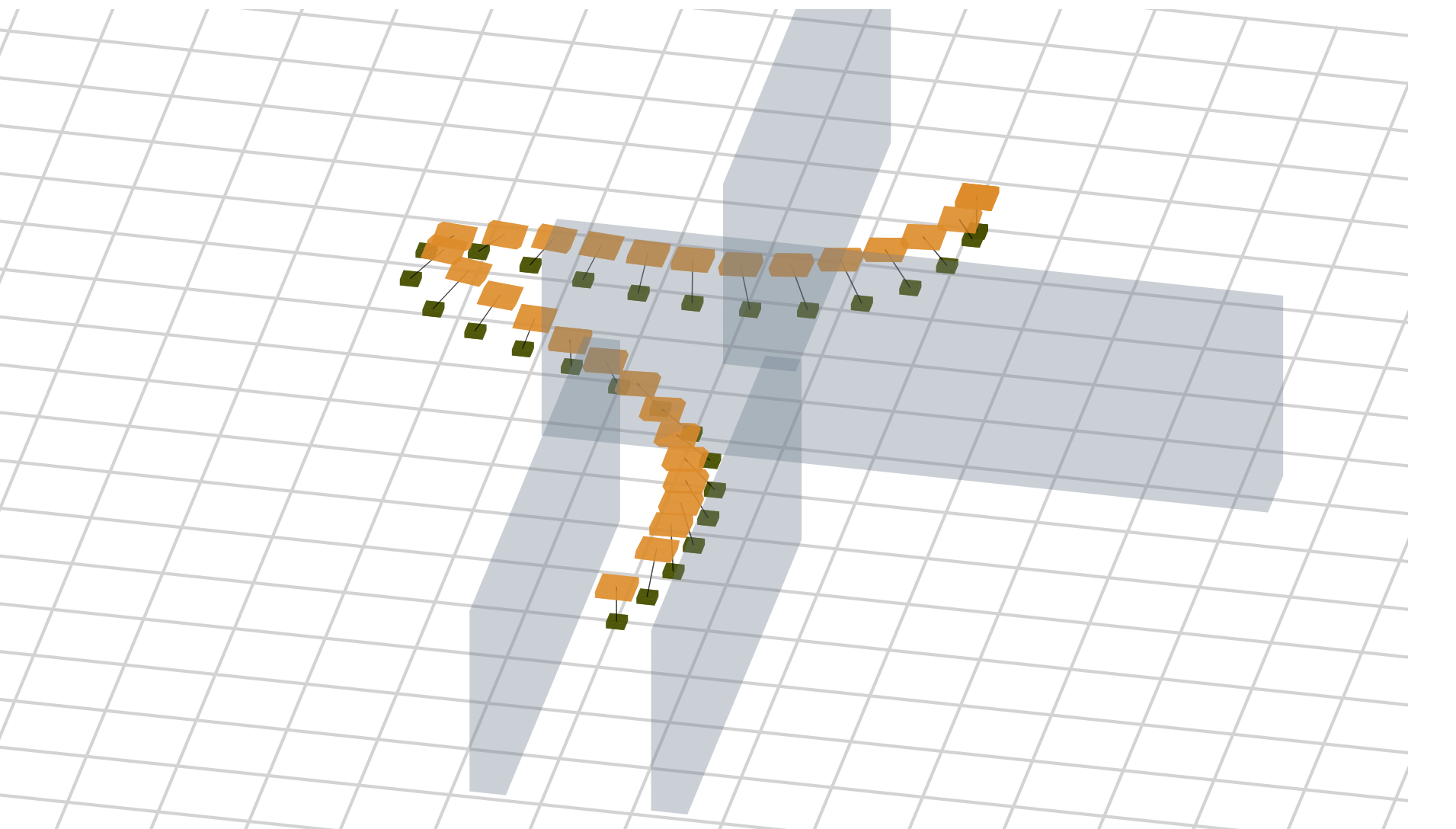}
  \end{minipage}\hfill%
  \begin{minipage}[b]{.24\textwidth}
    \includegraphics[width=\linewidth]{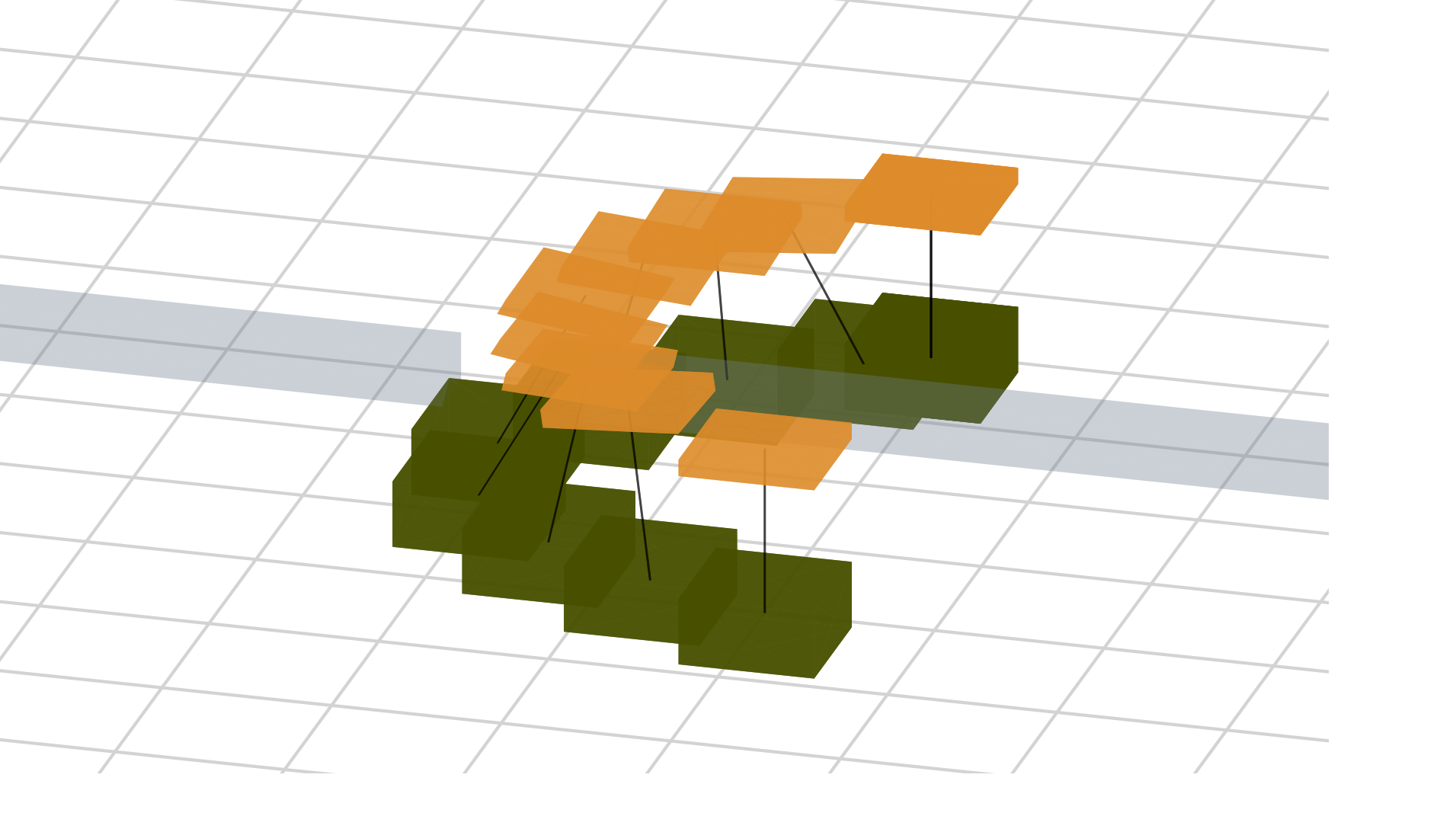}
  \end{minipage}\hfill%
  \begin{minipage}[b]{.24\textwidth}
    \includegraphics[width=\linewidth]{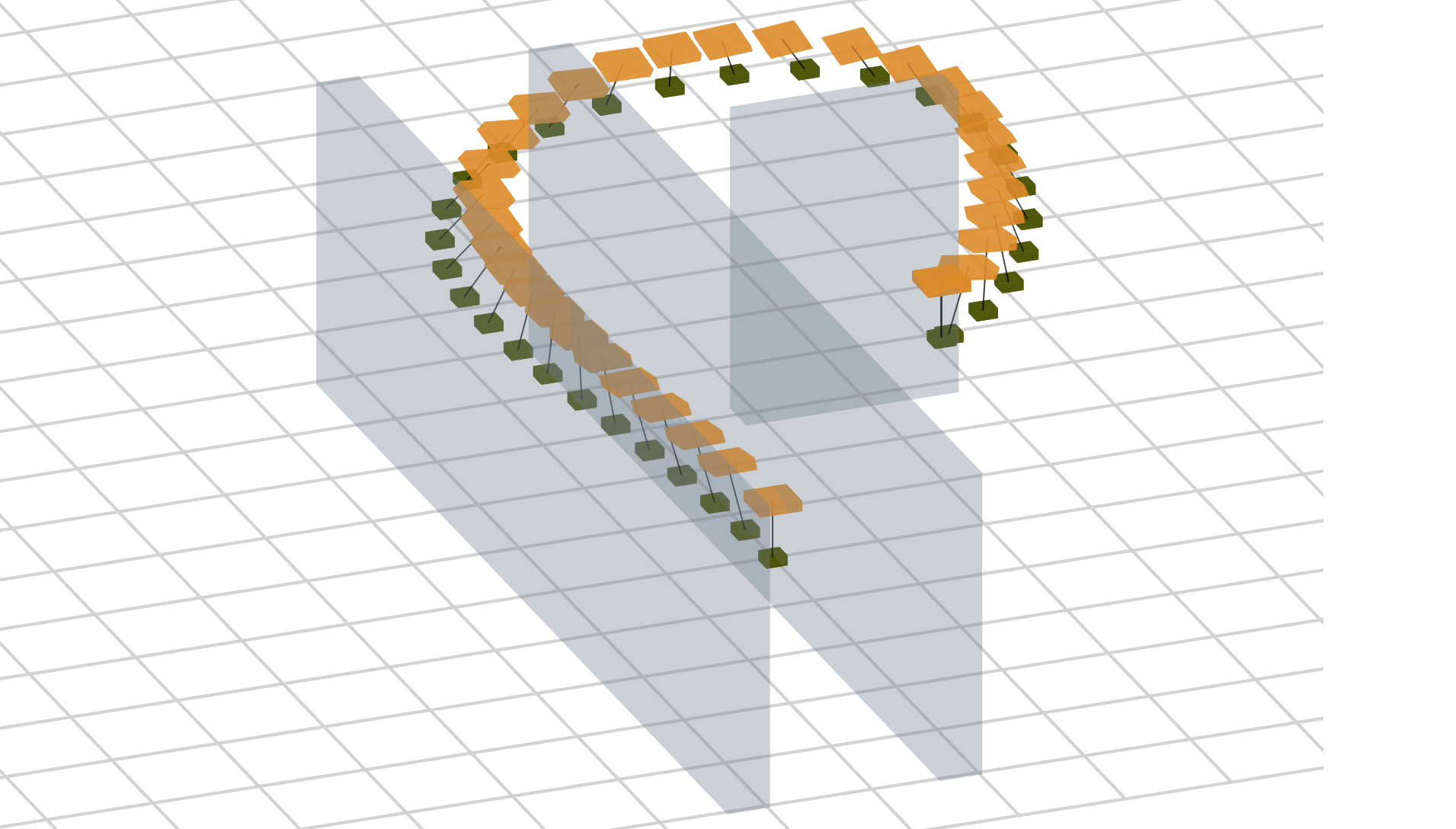}
  \end{minipage}

  \caption{Visualization of trajectories generated using PolyFly in 8 tightly cluttered environments. The \textcolor{customyellow}{quadrotor} is in yellow, \textcolor{customgreen}{payload} in green, \textcolor{customdarkgray}{cable} in gray and \textcolor{customlightblue}{obstacles} in light blue. Envs. 1-4 are shown in the top row and 5-8 in the bottom row.}
  \label{fig:traj_all_viz}
  \vspace{-10pt}
\end{figure*}

\begin{figure}[!t]          
  \centering
  \begin{minipage}[b]{.49\columnwidth}
    \includegraphics[width=\linewidth, trim=0 350 0 350, clip]{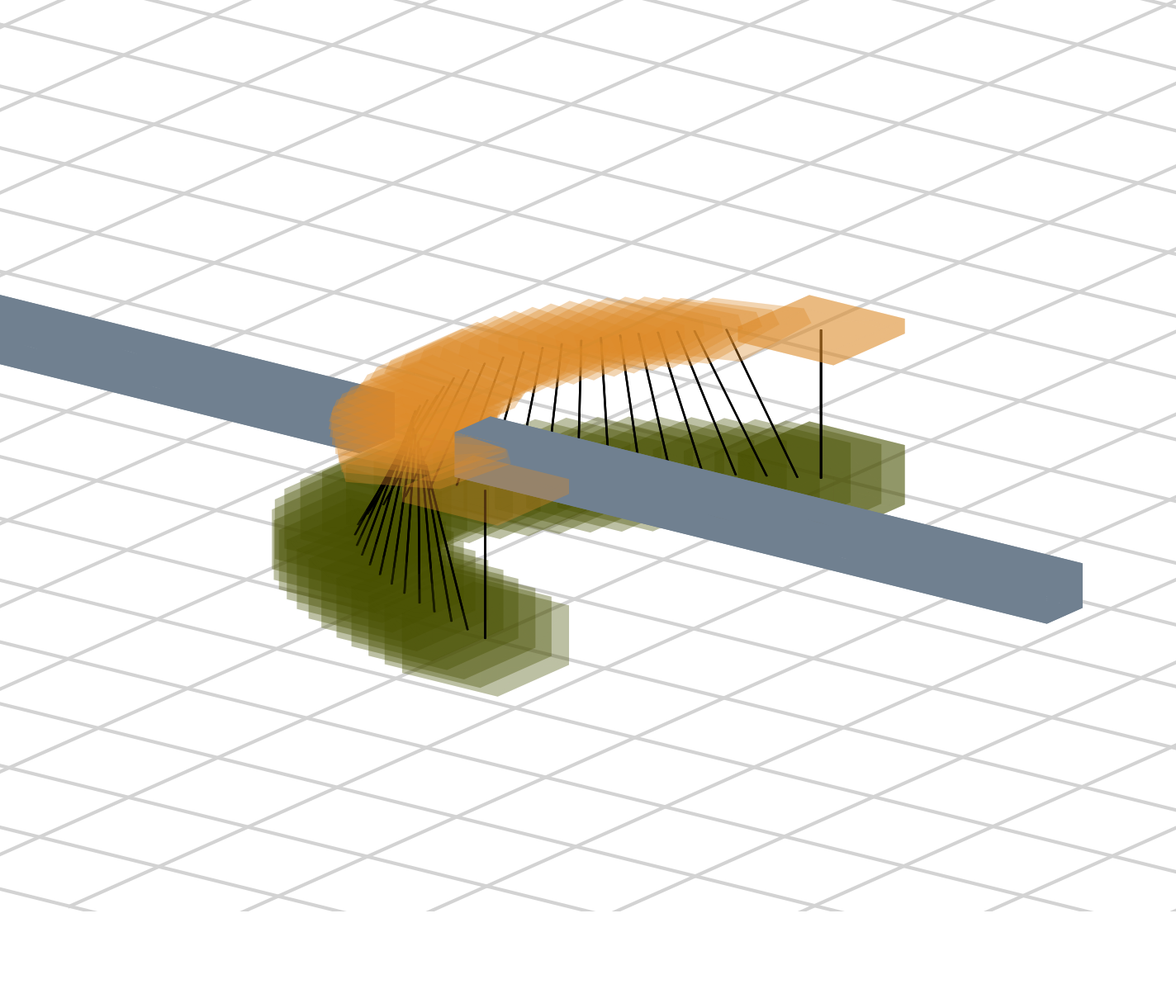}
  \end{minipage}\hfill%
  \begin{minipage}[b]{.49\columnwidth}
    \includegraphics[width=\linewidth, trim=150 450 200 300, clip]{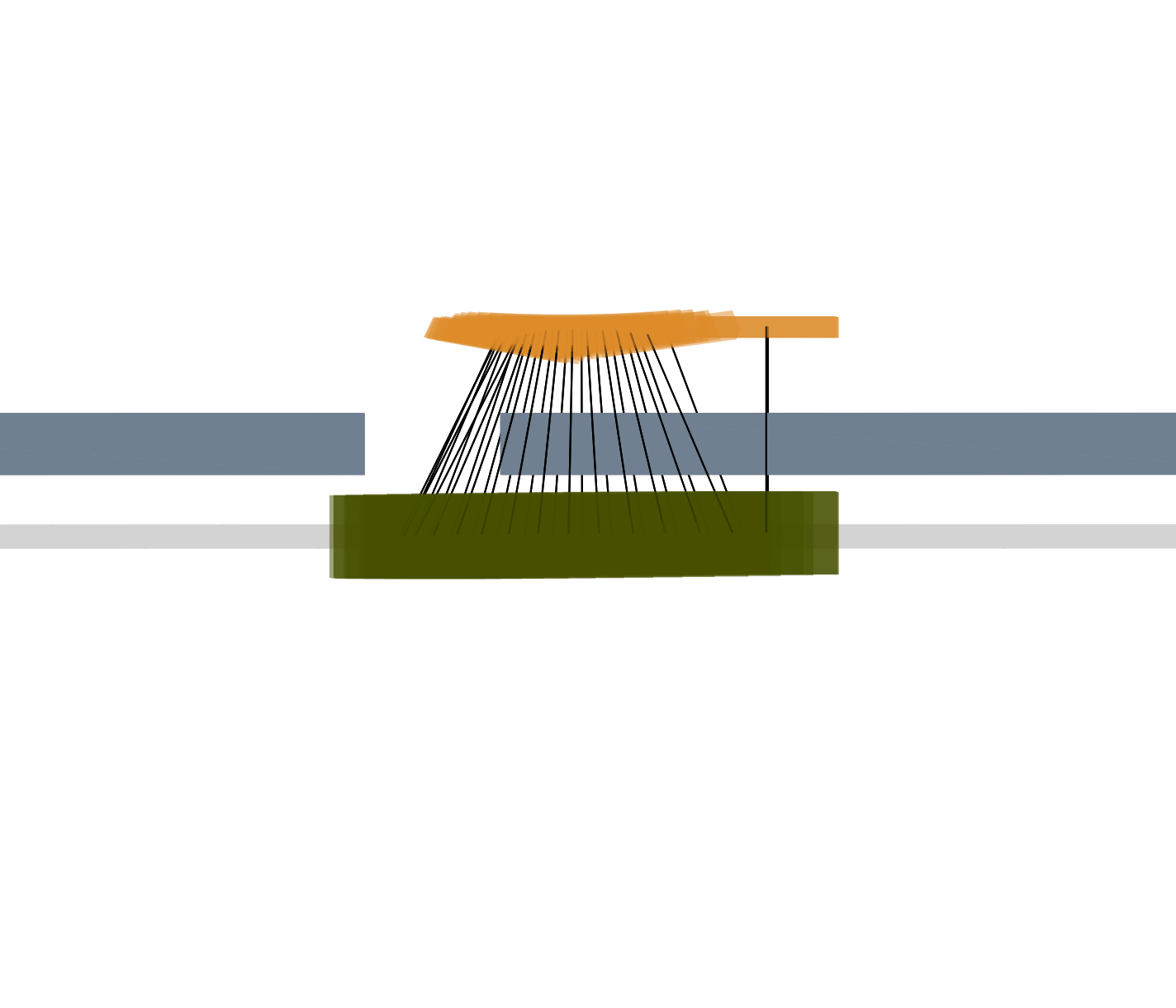}
  \end{minipage}
  \vspace{3pt}
  \begin{minipage}[]{.49\columnwidth}
    \includegraphics[width=\linewidth, trim=0 30 0 0, clip]{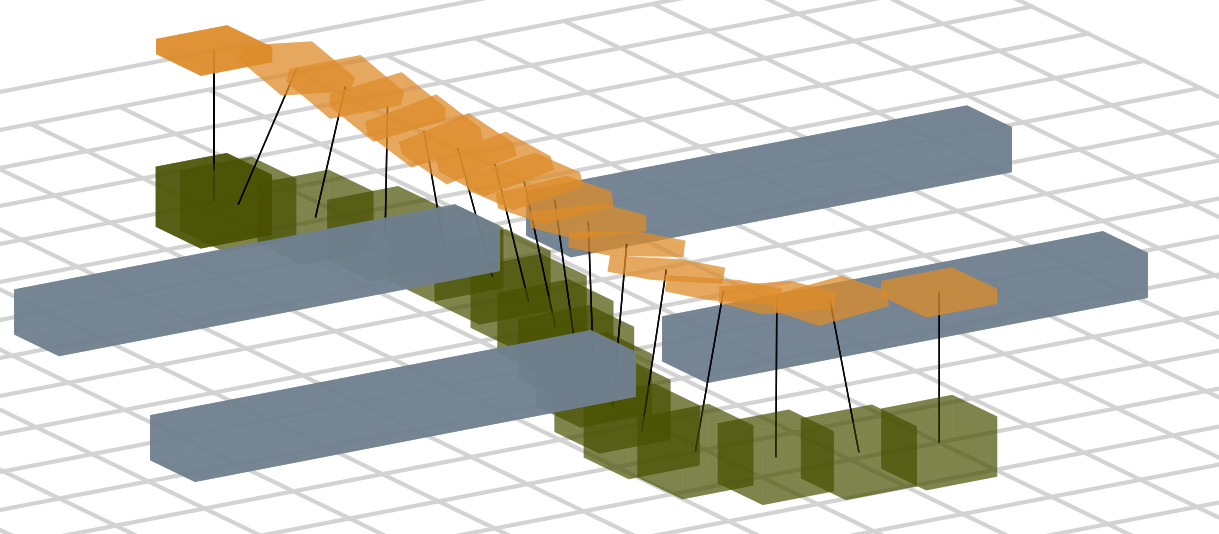}
  \end{minipage}\hfill%
  \begin{minipage}[]{.49\columnwidth}
    \includegraphics[width=\linewidth, trim=0 40 0 20, clip]{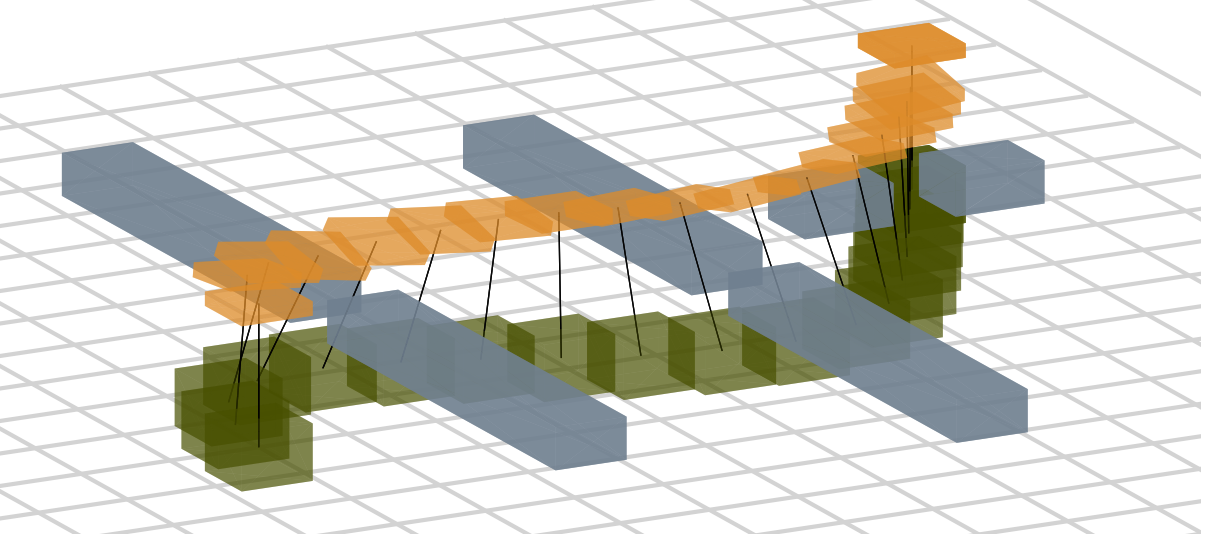}
  \end{minipage}
  \caption{PolyFly's trajectories for Env. 7 (top), 9 (bottom-left) and 10 (bottom-right) showcasing narrow-gap navigation.}
  \label{fig:traj_9_10}
      \vspace{-15pt}
\end{figure}

We conduct various experiments to evaluate the capabilities of this method. We seek to answer the following:

\begin{enumerate}
  \renewcommand\labelenumi{\textbf{Q\theenumi}} 
  \item Can this planning method produce trajectories that navigate tightly around corners of polytopic obstacles in cluttered maze environments?
  \item What are the benefits of modeling each component of the robot separately?
  \item What are the benefits of adding the quadrotor's orientation into the polytopic representation?
  \item Is adding a cost that regularizes the proximity to the initial guess important?
  \item Does our velocity initialization method help the solver converge?
  \item Can we run these aggressive trajectories on a real quadrotor with a suspended payload?
\end{enumerate}

For all experiments, we set $\Delta t_{min}=0.01 s$, $\Delta t_{max}=0.20s$, $\alpha_{u}=5.0$, $\alpha_{g}=5.0$, $\alpha_{td}=600.0$ and $N=100$. As described in Section \ref{subsec:q1}, $\alpha_{to}$ is tuned to maximize performance in each environment. 

\begingroup                              

\makeatletter                            
\renewcommand{\thesubsection}{Q\arabic{subsection}}        
\renewcommand{\thesubsectiondis}{\textbf{Q\arabic{subsection}}}

\renewcommand{\p@subsection}{\thesection-}

\makeatother
\setcounter{subsection}{0}               

\subsection{Navigating Cluttered Environments} \label{subsec:q1}
We create $10$ maze environments to evaluate the performance of our planning method. The optimization problem is constructed in CasADi \cite{Andersson2019} and solved with IPOPT \cite{Wachter2006ipopt}. 

We select AutoTrans, the MINCO approach \cite{wang2022minco} adopted in \cite{li2023autotrans, wang2024impact} as our primary baseline, which is an optimization-based trajectory generator that parametrizes trajectories as a sequence of $7^{th}$ order polynomials, initialized with kinodynamic hybrid $A^{*}$. They model the quadrotor and payload as independent spheres, the cable as a set of discrete points, and encode the environment using an ESDF. To solve the resulting problem, they reformulate the problem into an unconstrained one by applying $L_1$ penalties on the constraints.  

To ensure a fair comparison, we adjusted AutoTrans's safety margins, ESDF map resolution, kinodynamic search resolution, and quadrotor parameters to be consistent with our environment setup and removed AutoTrans' hardware limitations constraints on the quadrotor thrust and tilt angle. Furthermore, for each environment, we tune AutoTrans' and PolyFly's time minimization weights and report the mean results based on the planning output over $10$ trials in Table \ref{tab:topcat_baseline_times}. PolyFly's trajectories are shown in Fig. \ref{fig:traj_all_viz} and Fig. \ref{fig:traj_9_10}. 

We include an additional baseline, PolySingle, which mirrors our method but adopts the modeling of \cite{zheng2020}, where a single polytope represents the entire robot. 
AutoTrans fails to produce a collision-free trajectory in $5 / 10$ environments, as indicated by a {\textcolor{red}{\textbf{$\times$}}}. We also observed that in some cases, AutoTrans generated shorter paths in terms of distance, yet required more time to traverse them. This indicates that our method is able to better optimize for total flight time, even if that entails following a slightly longer path at higher speeds. Compared to PolySingle, our method successfully navigates the narrow gaps in Env. 7, 9 and 10, whereas PolySingle fails due to its modeling over-approximation. 
PolyFly's boost in performance compared to the AutoTrans may stem from the geometric modeling of the workspace and the use of the dual formulation, which produces smooth, non-conservative collision constraints \cite{zhang2020optimization}. These properties enable it to plan paths around obstacles while maintaining speed.
\definecolor{customgreen}{HTML}{6aae88}
\definecolor{darkgreen}{HTML}{156a0c}
\begin{table*}[t]
    \centering
    \renewcommand{\arraystretch}{1.15}
    \caption{PolyFly vs. PolySingle vs AutoTrans. Statistics are calculated after running 10 trials in each environment. $T$ is the mean trajectory duration in seconds, Path is mean path length in meters, Comp. is the mean optimization wall time in seconds, and SR is the success ratio. $\%$ $T$ \textcolor{darkgreen}{$\uparrow$} w/ $A$ denotes the $\%$ by which PolyFly’s trajectory duration $T$ is shorter than method $A$’s. {\textcolor{red}{\textbf{$\times$}}} marks a failure.}
    \label{tab:topcat_baseline_times}
    
    \begin{tabular*}{\textwidth}{@{\extracolsep{\fill}} l
      | c c c c c     
      | c c c c c     
      | c c c c       
      | c c @{}}
        \hline
        \hline
        & \multicolumn{5}{c|}{\textbf{PolyFly}} &
          \multicolumn{5}{c|}{\textbf{PolySingle (PS)}} &
          \multicolumn{4}{c|}{\textbf{AutoTrans (AT)}\cite{li2023autotrans}} &
          \multirow{2}{*}{\shortstack{\textbf{\% $T$ \textcolor{darkgreen}{$\uparrow$}}\\\textbf{w/ AT}}} &
          \multirow{2}{*}{\shortstack{\textbf{\% $T$ \textcolor{darkgreen}{$\uparrow$}}\\\textbf{w/ PS}}} \\
        \cline{2-6}\cline{7-11}\cline{12-15}
        \textbf{Env.} &
            \textbf{$T$} & \textbf{Path} & \textbf{Comp.} & \textbf{SR} & $\boldsymbol{\alpha_{to}}$ &
            \textbf{$T$} & \textbf{Path} & \textbf{Comp.} & \textbf{SR} & $\boldsymbol{\alpha_{to}}$ &
            \textbf{$T$} & \textbf{Path} & \textbf{Comp.} & \textbf{SR} &
            & \\ \hline
        1  & \textbf{3.90} & 4.51 & 4.83 & 1.0 & 5000 & 3.96 & 4.50 & 5.26 & 1.0 & 5000 & 4.10 & 4.45 & 0.06 & 1.0 & 4.9 & 1.5 \\
        2  & \textbf{4.60} & 6.24 & 3.97 & 1.0 & 6000 & 4.68 & 6.34 & 2.99 & 1.0 & 6000 & {\textcolor{red}{\textbf{$\times$}}} & {\textcolor{red}{\textbf{$\times$}}} & {\textcolor{red}{\textbf{$\times$}}} & 0.0 &
             \textbf{\textcolor{darkgreen}{$\infty$}} & 1.7 \\
        3  & \textbf{4.92} & 6.24 & 7.11 & 1.0 & 5000 & 5.13 & 6.35 & 7.51 & 1.0 & 5000 & 5.68 & 6.47 & 0.11 & 1.0 & 13.4 & 4.1 \\
        4  & \textbf{7.09} & 8.82 & 6.80 & 1.0 & 6000 & 7.22 & 9.16 & 5.16 & 1.0 & 6000 & 7.47 & 8.74 & 0.89 & 0.9 & 5.1 & 1.8 \\
        5  & \textbf{4.05} & 5.82 & 11.14 & 1.0 & 5000 & 4.12 & 5.82 & 11.20 & 1.0 & 5000 & 4.30 & 4.92 & 1.63 & 0.6 & 5.8 & 1.7 \\
        6  & \textbf{5.31} & 7.95 & 4.90 & 1.0 & 6000 & 5.48 & 8.08 & 7.10 & 1.0 & 6000 & 6.43 & 8.02 & 49.44 & 0.0 & 17.4 & 3.1 \\
        7  & \textbf{2.86} & 3.07 & 3.12 & 1.0 & 800  &
             {\textcolor{red}{\textbf{$\times$}}} & {\textcolor{red}{\textbf{$\times$}}} & {\textcolor{red}{\textbf{$\times$}}} & 0.0 & 800  &
             {\textcolor{red}{\textbf{$\times$}}} & {\textcolor{red}{\textbf{$\times$}}} & {\textcolor{red}{\textbf{$\times$}}} & 0.0 &
             \textbf{\textcolor{darkgreen}{$\infty$}} & \textbf{\textcolor{darkgreen}{$\infty$}} \\
        8  & \textbf{4.92} & 9.40 & 3.31 & 1.0 & 10000 & 4.96 & 9.51 & 3.21 & 1.0 & 10000 &
             {\textcolor{red}{\textbf{$\times$}}} & {\textcolor{red}{\textbf{$\times$}}} & {\textcolor{red}{\textbf{$\times$}}} & 0.0 &
             \textbf{\textcolor{darkgreen}{$\infty$}} & 0.8 \\
        9  & \textbf{3.72} & 5.89 & 3.34 & 1.0 & 2000 &
             {\textcolor{red}{\textbf{$\times$}}} & {\textcolor{red}{\textbf{$\times$}}} & {\textcolor{red}{\textbf{$\times$}}} & 0.0 & 2000 &
             {\textcolor{red}{\textbf{$\times$}}} & {\textcolor{red}{\textbf{$\times$}}} & {\textcolor{red}{\textbf{$\times$}}} & 0.0 &
             \textbf{\textcolor{darkgreen}{$\infty$}} & \textbf{\textcolor{darkgreen}{$\infty$}} \\
        10 & \textbf{4.46} & 6.89 & 9.92 & 1.0 & 2000 &
             {\textcolor{red}{\textbf{$\times$}}} & {\textcolor{red}{\textbf{$\times$}}} & {\textcolor{red}{\textbf{$\times$}}} & 0.0 & 2000 &
             {\textcolor{red}{\textbf{$\times$}}} & {\textcolor{red}{\textbf{$\times$}}} & {\textcolor{red}{\textbf{$\times$}}} & 0.0 &
             \textbf{\textcolor{darkgreen}{$\infty$}} & \textbf{\textcolor{darkgreen}{$\infty$}} \\
        \hline
        \hline
    \end{tabular*}
        \vspace{-15pt}
\end{table*}
\subsection{Independent Polytopic Models for Robot Components}
Modeling each robot component as a separate polytope allows the trajectory generator to accurately represent the system’s geometry. This fidelity is important when producing agile trajectories to navigate through narrow gaps and in tightly constrained environments where the robot must cut corners aggressively while still avoiding obstacles. 

The advantage is evident in Env. 7, 9 and 10, where the robot must pass through a narrow opening wider than the cable but thinner than the payload and quadrotor. Both the baselines fail, but our method successfully produces trajectories that thread the cable through the gaps, allowing the robot to safely navigate the environment. The real world experiment snapshots are shown in Fig. \ref{fig:cover_page}. 

PolySingle fails in these narrow-gap environments due to its single-polytope representation that over-approximates the robot’s geometry. On the other hand, AutoTrans models the cable as a discrete set of points and performs collision checks on each point, which should geometrically allow the cable to pass through the narrow gaps. Even though a solution exists, AutoTrans fails to produce a trajectory, even after extensive tuning of its optimization weights. Therefore, rather than the underlying geometry, the cause of the failures in AutoTrans may stem from other elements of the pipeline. First, the MINCO polynomial state parameterization, cannot, in general, represent optimal solutions \cite{wang2022minco} which limits its performance in tightly cluttered scenarios. And second, distance-field-based collision checking methods are not well-suited for applications that require accurate modeling of geometry \cite{schulman2013finding}. These methods compute the gradients to prevent collisions per point of the robot/obstacle, rather than the exact gradient needed to move the polytope out of collision. These two gradients do not always agree, and can cause issues when the robot travels in close proximity to obstacles. 

\subsection{Incorporating the Quadrotor's Orientation Into Polytopic Representation}
Accurately capturing the quadrotor’s orientation further increases the fidelity of the polytopic model and therefore navigation reliability. This detail becomes critical when the vehicle flies close to a ceiling or floor, particularly for a platform with long rotor arms that form a thin yet wide rectangular footprint. Since high-acceleration maneuvers induce large orientation changes, aggressive trajectories close to surfaces can cause rotor collisions with the environment. As shown in Fig. \ref{fig:poly} and \ref{fig:ceiling_examples}, most planners over-approximate the vehicle with a sphere or ellipsoid. By employing a tight polytope that accounts for the robot's orientation, we eliminate this conservatism and ensure collision free navigation in near-ceiling flight.

We demonstrate these benefits in Fig. \ref{fig:incorporating_orientation} and \ref{fig:ceiling_examples}. Fig. \ref{fig:incorporating_orientation} illustrates how incorporating the orientation into the planner prevents collisions. In Fig. \ref{fig:incorporating_orientation} (A), the quadrotor strikes the ceiling during a high-acceleration maneuver because the planner ignores the orientation change that accompanies acceleration. In contrast, Fig. \ref{fig:incorporating_orientation} (B) shows how the orientation-aware planner descends during the acceleration phase to avoid a collision. The ceiling-floor scenario in Fig. \ref{fig:ceiling_examples} shows the PolyFly trajectory and overlays the PolySingle polytope to illustrate the over-approximation introduced by \cite{zheng2020}. PolySingle fails in these scenarios because \cite{zheng2020} uses a common orientation for all components due to its single polytope representation of the entire system, which prevents navigation in such tightly constrained environments.

\begin{figure}[!t]          
  \centering
  \begin{minipage}[b]{.48\columnwidth}
    \includegraphics[width=\linewidth, trim=0 0 100 50, clip]{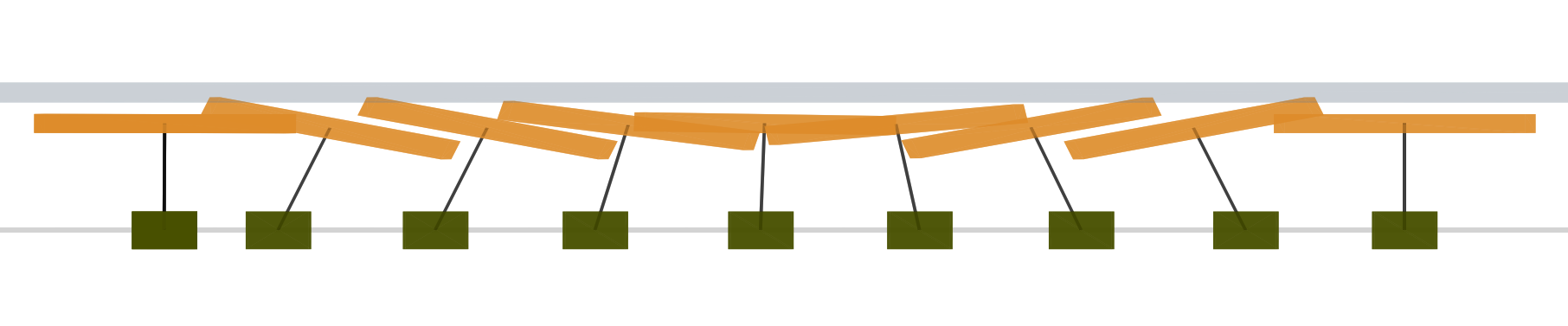}
    \captionof*{figure}{(A) Robot Collides}
  \end{minipage}\hfill%
  \begin{minipage}[b]{.48\columnwidth}
    \includegraphics[width=\linewidth,  trim=0 0 0 0, clip]{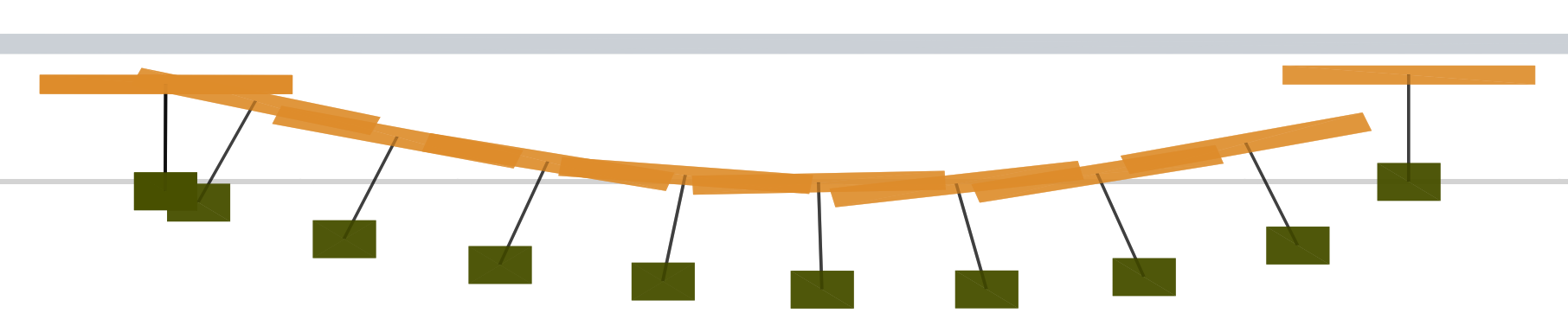}
    \captionof*{figure}{(B) Robot Avoids Collision}
  \end{minipage}
  \caption{Left: An orientation-agnostic planner produces a trajectory that collides with the ceiling. Right: By accounting for the robot's orientation, the optimized path descends to prevent collisions while accelerating.}
  \label{fig:incorporating_orientation}  
  \vspace{5pt}
  \centering
  \begin{minipage}[b]{.49\columnwidth}
    \includegraphics[width=\linewidth, trim=0 0 0 0, clip]{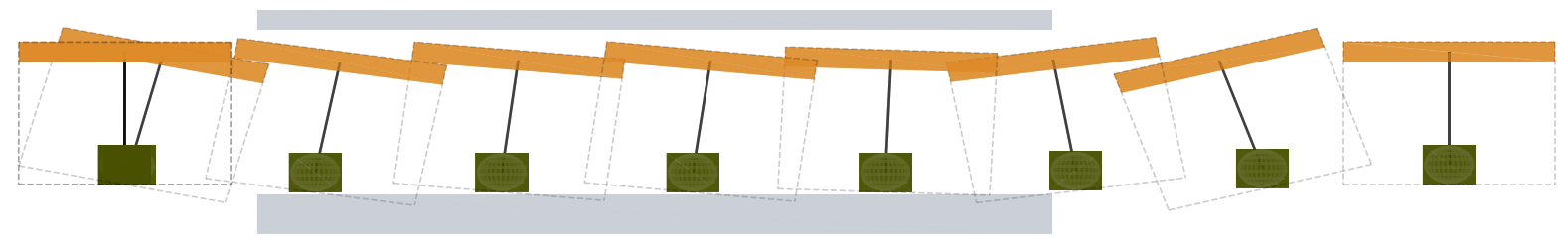}
  \end{minipage}\hfill%
  \begin{minipage}[b]{.49\columnwidth}
    \includegraphics[width=\linewidth, trim=0 0 0 0, clip]{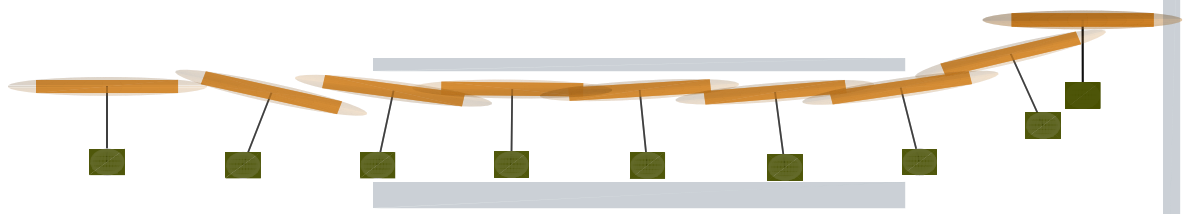}
  \end{minipage}
  \caption{Optimized trajectories by PolyFly for the ceiling-floor (left) and ceiling-wall (right) environments. The dashed lines are the polytopic approximation by PolySingle and the semi-transparent ellipse is the ellipsoidal approximation.
  }
  \label{fig:ceiling_examples}
  \vspace{-10pt}
\end{figure}

\subsection{Regularizing Proximity to Initial Guess} \label{subsec:prox_guess}
We perform an ablation study to determine the importance of regularizing the proximity to the initial guess. We run the optimizer with $3$ different $\alpha_g$ values. The results are shown in Table 
\ref{tab:alpha_g_ablation}. PolyFly fails in $5/10$ environments if $\alpha_g=0$. Trajectories generated with $\alpha_g=1$ are on average slower than those generated with $\alpha_g=5$. This highlights the value of this regularization term. We believe this aids convergence by incentivizing the optimizer to remain in the vicinity of the initial guess, preventing it from entering regions of local minima that are difficult to escape from.

\definecolor{customgreen}{HTML}{6aae88}
\definecolor{darkgreen}{HTML}{156a0c}
\begin{table}[!t]
    \centering
    \setlength{\tabcolsep}{7pt}
    \renewcommand{\arraystretch}{1.15}
    \caption{$T$ for different values of $\alpha_g$. \textbf{\% $T$ \textcolor{darkgreen}{$\uparrow$}~(5 w/ 1)} indicates improvement of $\alpha_g=5$ over $\alpha_g=1$. 
    }
    \label{tab:alpha_g_ablation}
    \begin{tabular}{lcccc}
        \hline
        \hline
        \textbf{Env.} & $\alpha_g=0$ & $\alpha_g=1$ & $\alpha_g=5$ & \textbf{\% $T$ \textcolor{darkgreen}{$\uparrow$}~(5 w/ 1)} \\ \hline
        1  & {\textcolor{red}{\textbf{$\times$}}} & 3.92 & 3.92 & 0.0\% \\
        2  & {\textcolor{red}{\textbf{$\times$}}} & 5.02 & 4.61 & 8.2\% \\
        3  & 4.94 & 4.96 & 4.97 & -0.2\% \\
        4  & {\textcolor{red}{\textbf{$\times$}}} & 7.13 & 7.15 & 0.3\% \\
        5  & 4.20 & 4.03 & 4.05 & -0.5\% \\
        6  & {\textcolor{red}{\textbf{$\times$}}} & 5.48 & 5.37 & 2.0\% \\
        7  & 3.15 & 4.59 & 3.16 & 31.2\% \\
        8  & {\textcolor{red}{\textbf{$\times$}}} & 4.97 & 4.99 & -0.4\% \\
        9  & 3.93 & 3.93 & 3.94 & -0.3\% \\
        10 & 4.67 & 4.72 & 4.69 & 0.6\% \\ 
        \hline
        \hline
    \end{tabular}
        \vspace{-5pt}
\end{table}
\subsection{Payload Velocity Initialization}
\definecolor{customgreen}{HTML}{6aae88}
\definecolor{darkgreen}{HTML}{156a0c}
\begin{table}[!t]
    \centering
    \setlength{\tabcolsep}{7pt}
    \renewcommand{\arraystretch}{1.15}
    \caption{Performance with and without $\loadvel$ initialization.
             }
    \label{tab:vel_init_comparison}

    \begin{tabular}{lcccc}
        \hline
        \hline
        & \multicolumn{2}{c}{\textbf{With Velocity Initialization}} &
          \multicolumn{2}{c}{\textbf{W/o Velocity Initialization}} \\
        \cmidrule(lr){2-3}\cmidrule(lr){4-5}
        \textbf{Env.} & $T$ (s) & \textbf{Iterations} &
                       $T$ (s) & \textbf{Iterations} \\
        \hline
        1  & 3.90 & 201\,\textcolor{darkgreen}{$\downarrow$} & 3.98 & 745 \\
        2  & 4.60 & 223\,\textcolor{darkgreen}{$\downarrow$} & 4.61 & 640 \\
        3  & 4.92 & 133\,\textcolor{darkgreen}{$\downarrow$} & 4.92 & 677 \\
        4  & 7.09 & 185\,\textcolor{darkgreen}{$\downarrow$} & {\textcolor{red}{\textbf{$\times$}}} & {\textcolor{red}{\textbf{$\times$}}} \\
        5  & 4.05 & 353\,\textcolor{darkgreen}{$\downarrow$} & {\textcolor{red}{\textbf{$\times$}}} & {\textcolor{red}{\textbf{$\times$}}} \\
        6  & 5.31 & 206\,\textcolor{darkgreen}{$\downarrow$} & 5.31 & 559 \\
        7  & 2.86 & 234\,\textcolor{darkgreen}{$\downarrow$} & 2.87 & 309 \\
        8  & 4.92 & 161\,\textcolor{darkgreen}{$\downarrow$} & 4.90 & 617 \\
        9  & 3.72 & 112\,\textcolor{darkgreen}{$\downarrow$} & 3.72 & 206 \\
        10 & 4.46 & 306\,\textcolor{darkgreen}{$\downarrow$} & 4.43 & 652 \\
        \hline
        \hline
    \end{tabular}
\vspace{-5pt}
\end{table}
To evaluate the effectiveness of our solver initialization strategy, we conduct an ablation study that compares the payload-velocity initialization of eq.~(\ref{eq:vel_init}) against a variant where velocities are initialized to zero. Table \ref{tab:vel_init_comparison} summarizes the results.
The solver fails in $2/10$ environments without the velocity initialization strategy. For all other scenarios, our strategy reduces the solver iterations by an average of $59\%$.

\subsection{Real World Experiments}
We run real world experiments with Env. $1$, $3$ and $7$ to validate that our method produces trajectories that can be deployed on real robots. Our system architecture is shown in Fig. \ref{fig:system_architecture}. For real-world experiments only, we use the Non-Linear Model Predictive Control (NMPC) setup from \cite{sarvaiya2024hpa} as our controller. The NMPC uses the dynamics eqs.~\eqref{eq:single-kinematics}--\eqref{eq:thrust_map} with inputs $f_{1-4}$. After generating a trajectory with PolyFly, we send the desired $\mathbf{\tau}$ to the NMPC at $200$ Hz. As in \cite{sarvaiya2024hpa}, the predicted collective thrust and bodyrates from the MPC's second stage are sent to the low-level controller. The robot's states are measured using a motion capture system and are upsampled using Kalman Filters. Snapshots of our experiments are shown in Fig. \ref{fig:cover_page}. The position errors for the quadrotor and payload are shown in Fig. \ref{fig:tracking_errors} and Table \ref{tab:axis_error_table}. Our results indicate that PolyFly produces dynamically feasible trajectories that can be tracked by an onboard controller with low errors. The supplementary video contains real-world footage that highlights the agile trajectories generated by our method.
\begin{figure}[!t]
    \includegraphics[width=\columnwidth, trim=170 270 260 110, clip]{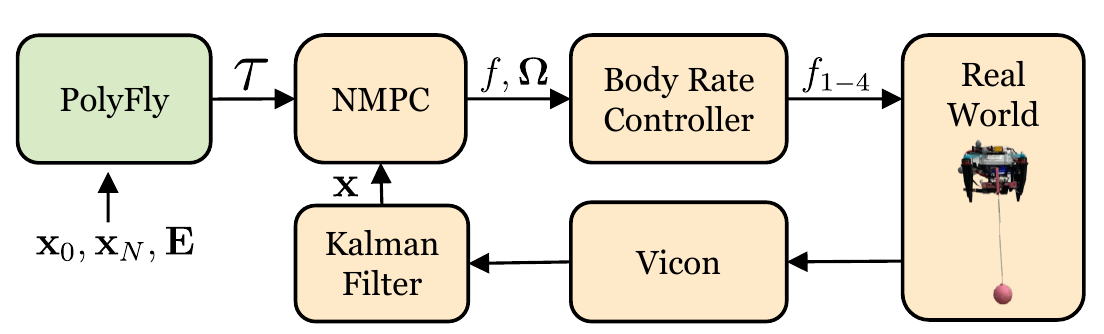}
    \caption{System architecture for real-world deployment.}
    \label{fig:system_architecture}
    \vspace{-10pt}
\end{figure}
\begin{figure}[!t]   
    \centering
    \includegraphics[width=\columnwidth, trim=20 60 0 0, clip]{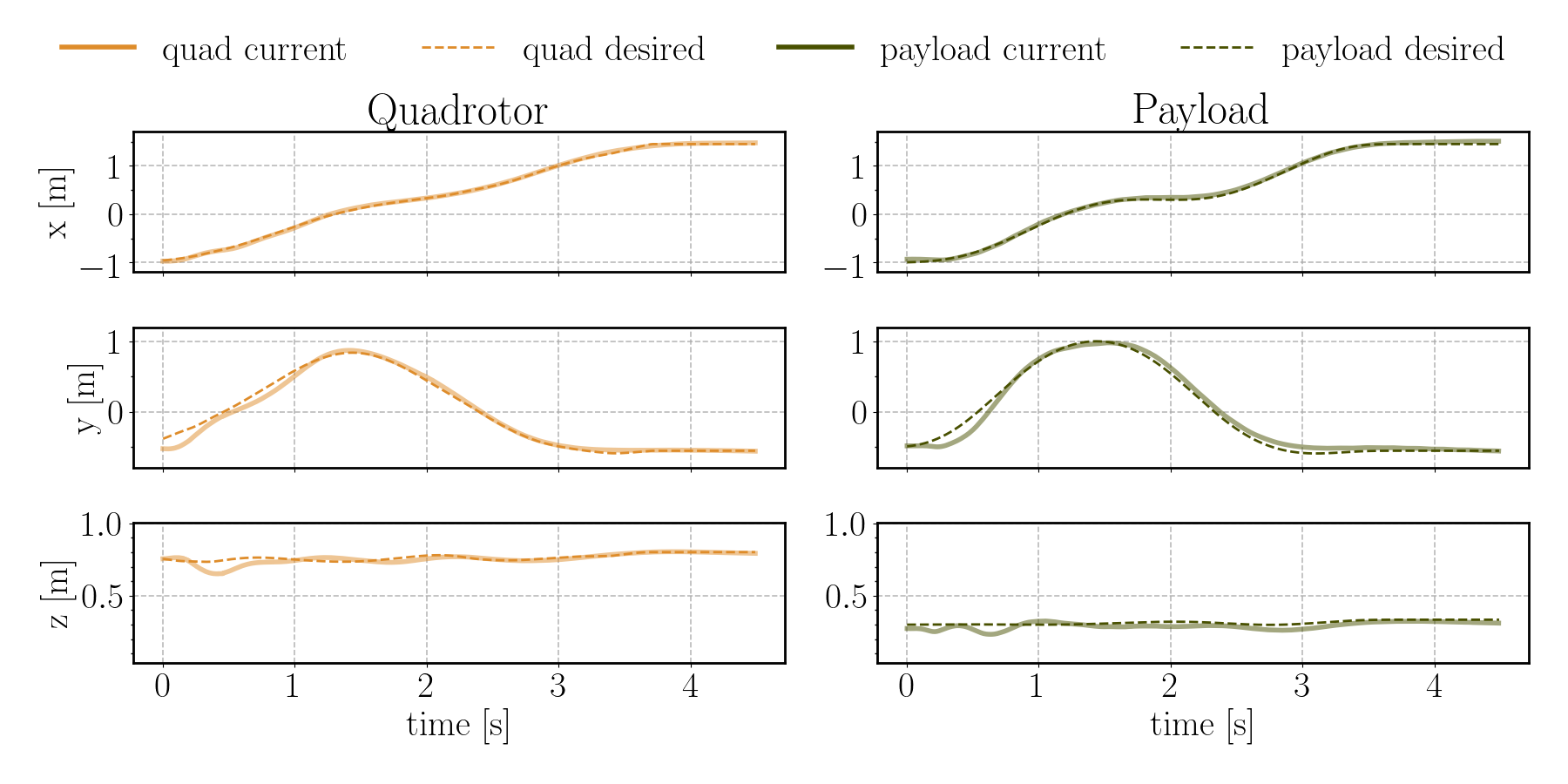}\\[-0.4em]
    \vspace{3pt}
    \includegraphics[width=\columnwidth, trim=20 60 0 70, clip]{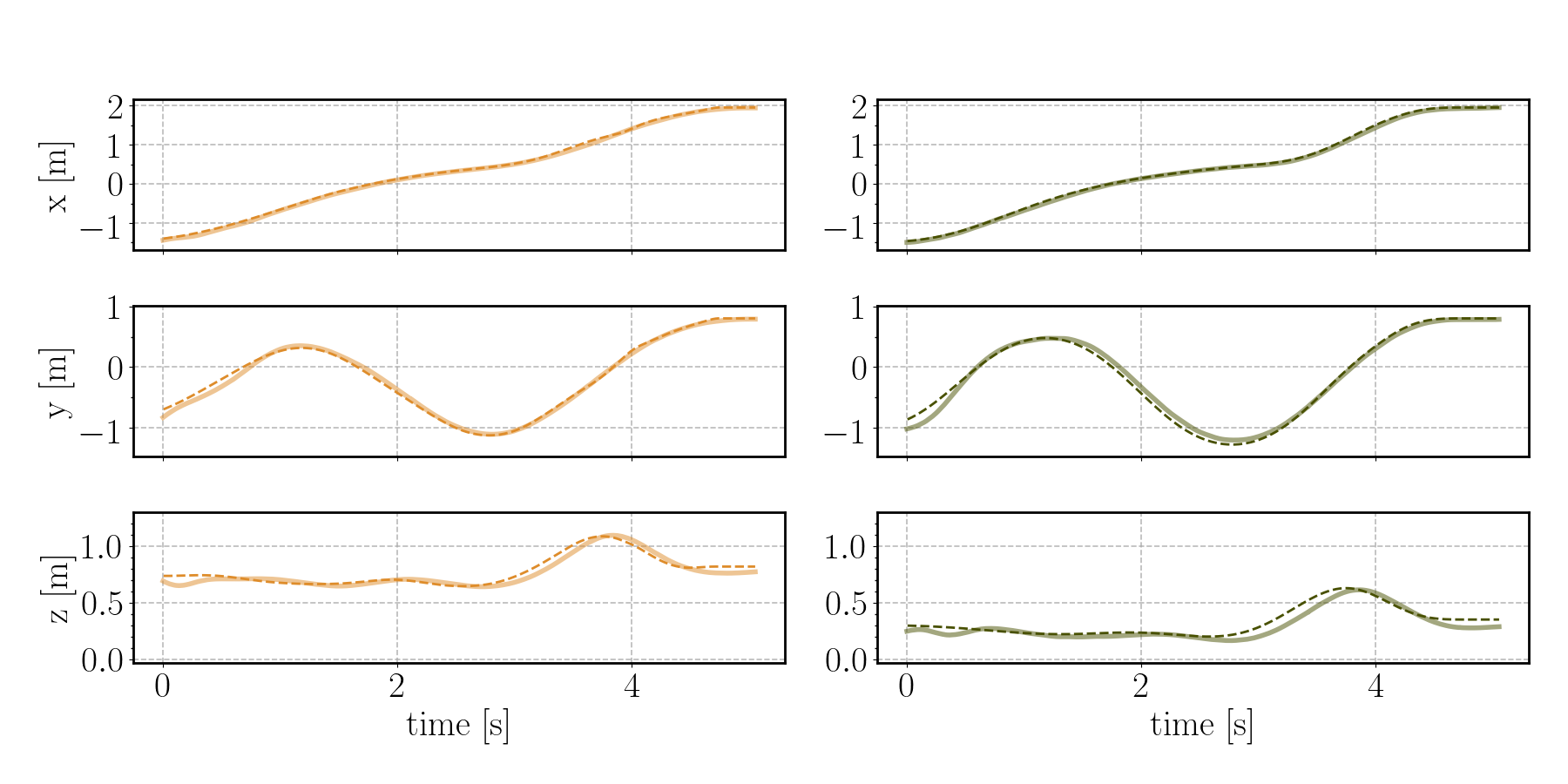}\\[-0.4em]
    \vspace{3pt}
    \includegraphics[width=\columnwidth, trim=20 0 0 70, clip]{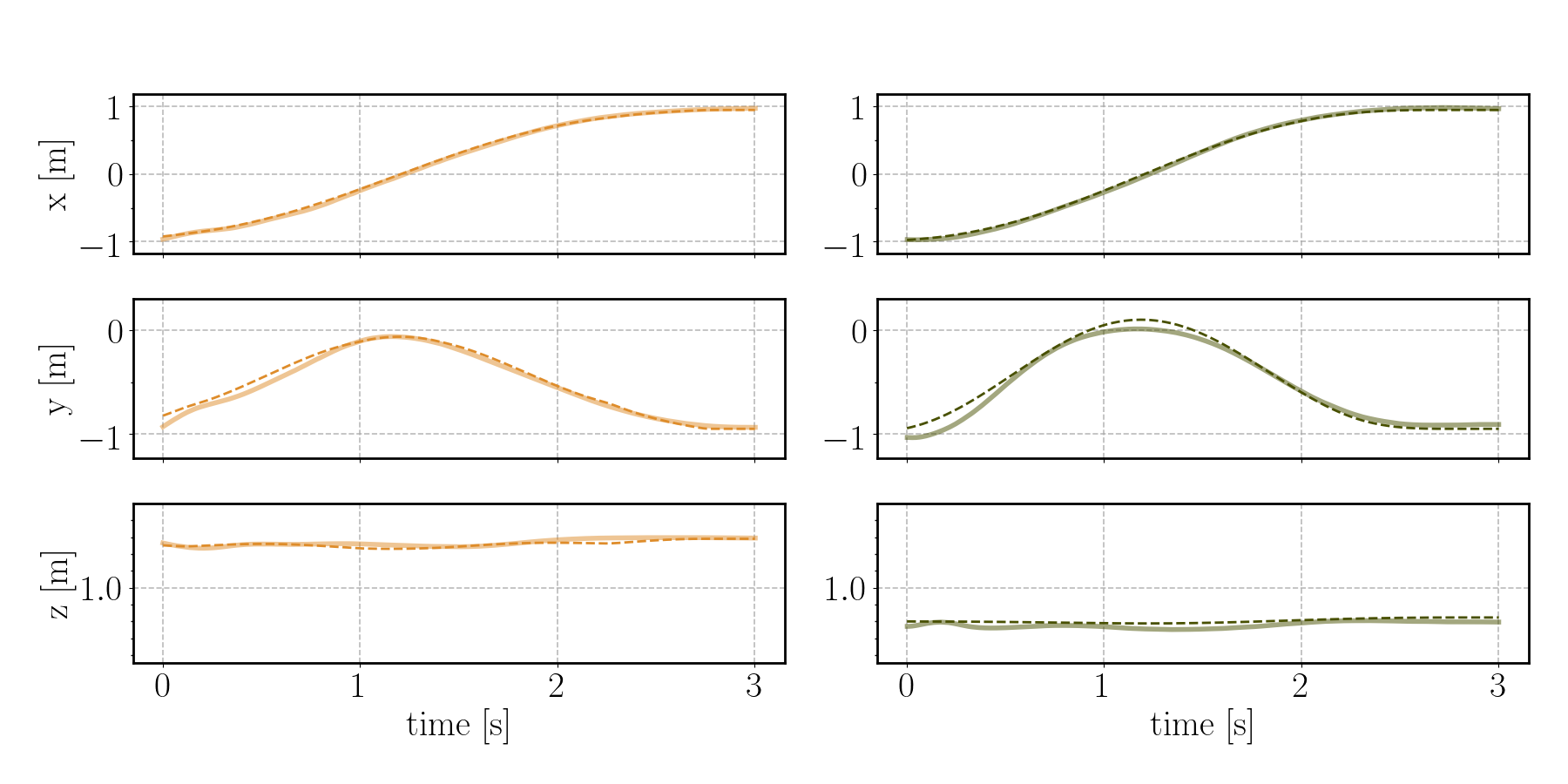}\\[-0.4em]
    \vspace{2pt}
    \caption{Desired vs. Measured positions of the quadrotor and payload during real-world experiments for Env. 1, 3, and 7.}
    \label{fig:tracking_errors}
    \vspace{-10pt}
\end{figure}


\begin{table}[!t]
    \centering
    \setlength{\tabcolsep}{8pt}
    \renewcommand{\arraystretch}{1.2}
    \caption{Tracking error statistics (meters)}
    \label{tab:axis_error_table}

    \begin{tabular}{c c c c c c c}
        \hline
        \hline
        \multirow{2}{*}{\textbf{Axis}} & \multicolumn{2}{c}{\textbf{Env. 1}} & \multicolumn{2}{c}{\textbf{Env. 3}} & \multicolumn{2}{c}{\textbf{Env. 7}} \\
        & \textbf{Mean} & \textbf{Std} & \textbf{Mean} & \textbf{Std} & \textbf{Mean} & \textbf{Std} \\
        \hline
        $x$ & 0.027 & 0.017 & 0.029 & 0.016 & 0.021 & 0.094 \\
        $y$ & 0.058 & 0.047 & 0.059 & 0.048 & 0.052 & 0.036 \\
        $z$ & 0.024 & 0.012 & 0.039 & 0.029 & 0.029 & 0.087 \\
        \hline
        \hline
    \end{tabular}
        \vspace{-15pt}
\end{table}
\section{Conclusion} \label{sec:conclusion}
This paper presented PolyFly, a global planner for a single robot aerial transportation system that models the robot and environment using a set of polytopes and enables the generation of collision-free trajectories in cluttered scenarios. Our proposed method yielded consistently faster trajectories than a popular state-of-the-art baseline in ten different maze-like environments. Experimental results confirm the trajectories’ dynamic feasibility and the applicability of the proposed solution in real-world settings. One limitation is that PolyFly does not yet run at real-time frequencies or account for onboard obstacle sensing. Future work will address this by efficiently updating trajectories using onboard local obstacle information and leveraging advances in polytopic collision-avoidance checks \cite{wu2024} to reduce computation.

\bibliographystyle{IEEEtran}
\bibliography{main}

\end{document}